\documentclass[10pt,twocolumn,letterpaper]{article}

\usepackage[pagenumbers]{cvpr} 

\usepackage{graphicx}
\usepackage{amsmath}
\usepackage{amssymb}
\usepackage{booktabs}
\usepackage{multirow}
\usepackage{xcolor}
\usepackage{color, colortbl}
\usepackage{bm}
\usepackage{enumitem}

\definecolor{tb_bg_color}{rgb}{0.835, 0.902, 0.941}

\newcommand{\mypara}[1]{\vspace{1mm}\noindent\textbf{#1}}

\usepackage[pagebackref,breaklinks,colorlinks]{hyperref}

\usepackage[capitalize]{cleveref}
\crefname{section}{Sec.}{Secs.}
\Crefname{section}{Section}{Sections}
\Crefname{table}{Table}{Tables}
\crefname{table}{Tab.}{Tabs.}

\def\cvprPaperID{5773} 
\def\confName{CVPR}
\def\confYear{2023}

\begin{document}

\title{$\alpha$ DARTS Once More: Enhancing Differentiable Architecture Search \\
by Masked Image Modeling}

\author{Bicheng Guo\textsuperscript{1}\thanks{Part of the work was done during an internship at Alibaba Group.}, Shuxuan Guo\textsuperscript{3}, Miaojing Shi\textsuperscript{4}, Peng Chen\textsuperscript{2}, Shibo He\textsuperscript{1}\thanks{Corresponding author}~, Jiming Chen\textsuperscript{1}, Kaicheng Yu\textsuperscript{2}
\\
$^1$Zhejiang University, $^2$Alibaba Group, $^3$CVLab, EPFL,  $^4$King's College London
\\
{\tt \small \{guobc, s18he, cjm\}@zju.edu.cn} \quad {\tt \small shuxuan.guo@epfl.ch}\\
{\tt \small miaojing.shi@kcl.ac.uk} \quad {\tt\small\{yuanshang.cp, yukaicheng.ykc\}@alibaba-inc.com} }

\maketitle

\begin{abstract}
Differentiable architecture search~(DARTS) has been a mainstream direction in automatic machine learning. 
Since the discovery that original DARTS will inevitably converge to poor architectures, recent works alleviate this by either designing rule-based architecture selection techniques or incorporating complex regularization techniques, 
abandoning the simplicity of the original DARTS that selects architectures based on the largest parametric value, namely $\alpha$. 
Moreover, we find that all the previous attempts only rely on classification labels, hence learning only single modal information and limiting the representation power of the shared network. 
To this end, we propose to additionally inject semantic information by formulating a patch recovery approach. Specifically, we exploit the recent trending masked image modeling and do not abandon the guidance from the downstream tasks during the search phase.  
Our method surpasses all previous DARTS variants and achieves the state-of-the-art results on CIFAR-10, CIFAR-100 and ImageNet without complex manual-designed strategies.
\end{abstract}
\vspace{-8pt}
\section{Introduction}
\label{sec:intro}
\vspace{-4pt}
Neural architecture search~(NAS) has been a de facto standard to automatically design a neural network.\cite{Elsken_2019_survey, zoph2017neural,Zoph_2018_nasnet, real_2017_largeevolution, liu2018hierarchical},
where the differentiable architecture search~(DARTS)~\cite{liu2018darts,cai2018proxylessnas,Chen_2019_pdarts,ye_2022_beta,Zela2020Understanding} plays a major role in NAS domain.
In essence, instead of using a one-hot encoding to select the architecture out of a weight-sharing network~\cite{Guo_2020_spos, pham18ENAS, Chu_2021_fairnas,Cai2020Once-for-All},
DARTS-based approaches exploit a continuous relaxation of the architecture, \ie, using an architecture parameter, usually dubbed as $\alpha$, to compute the probability of operations selection. 
By formulating a bi-level optimization, DARTS could back-propagate the loss directly to the $\alpha$.

However, there exist several downsides of the differentiable architecture search, of which Yu \etal identify that the original DARTS cannot surpass the random search~\cite{Yu2020Evaluating}.
Zela \etal further identify a performance collapse issue in that the edges of the learned architectures are dominated by the parameter-free operations, \eg, \textit{skip} connections~\cite{Zela2020Understanding}. 
To this end, various approaches have been proposed, such as regularizing on the empirical indicators~\cite{Zela2020Understanding};
explicitly limiting the number of the \textit{skip} connections~\cite{Chen_2019_pdarts,liang2019darts+};
perturbing and smoothing the precipitous validation loss landscape~\cite{chen_2020_sdarts}; 
factoring out the unfair advantages of \textit{skip} connections~\cite{chu2021dartsminus,chu_2020_fairdarts}.
After Wang \etal ~\cite{wang_2021_rethinking} discover that $\alpha$ cannot truthfully represent the architecture strength and is the main cause of instability of DARTS,
Ye \etal~ propose to regularize the probability vector, \ie, applying softmax function over $\alpha$ 
and achieves state-of-the-art performance.
While interesting, these approaches either feature 
hand-crafted design to explicitly avoid \textit{skip} connection, or involve extra 
procedures
to complicate the search phase.

\begin{figure}[t]
    \centering
     \includegraphics[width=1\linewidth]{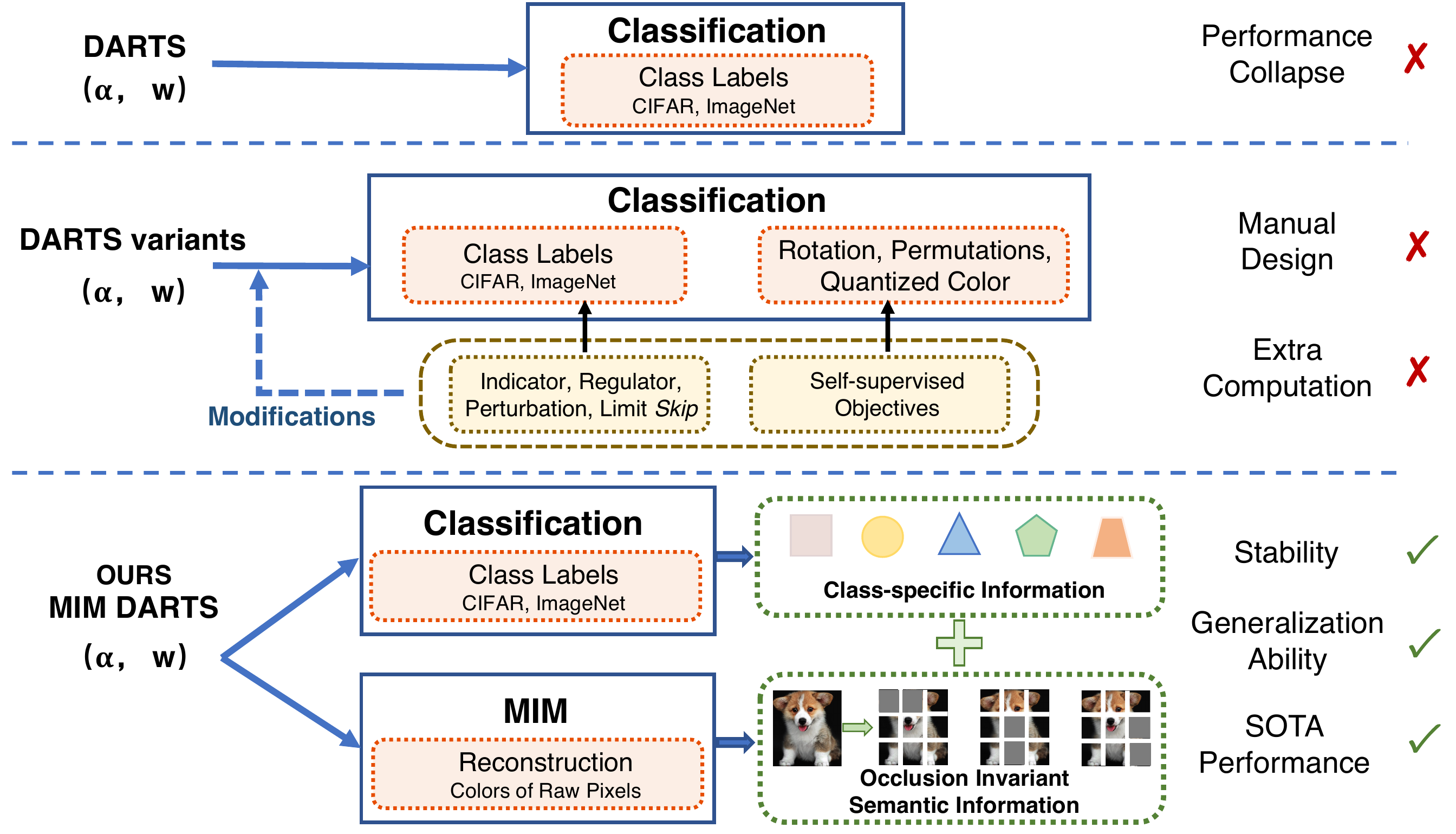}
     \vspace{-6pt}
     \caption{\textbf{Comparison of the original DARTS, DARTS variants, and our MIM-DARTS.}
     Our framework steps out of the traditional differentiable architecture search which performs only the classification task
     and embraces the occlusion invariant semantic information learned from the MIM task.
     }
     \label{fig:fig1}
     \vspace{-8pt}
  \end{figure}

A recent study shows that insufficient training of original DARTS favors the architectures that converge quickly, which are usually wide and shallow~\cite{shu_2020_understanding}. 
We argue that the key to all these 
aforementioned
approaches is that they intentionally slow down the training process by various regularizations of altering the search pipeline. 
We seek an alternative approach that injects more information into the optimization to avoid fast convergence, without introducing extra annotations. 
To this end, we propose a simple yet effective 
framework, that tempts to let the shared network learn occlusion invariant semantic information~\cite{Kong_2022_Occlusion} in addition to the traditional classification labels.
Specifically, apart from the original optimization target, we exploit the Masked Image Modeling~(MIM)~\cite{bao2022beit,He_2022_mae}
task to mask and recover random patches during the training. 
We dub our method MIM-DARTS. 
In contrast to those previous methods, ours does not incorporate any manually designed regularization and preserves the original DARTS search pipeline, to make the original $\alpha$ DARTS yesterday once more, as shown in~\cref{fig:fig1}.

There has been a preliminary attempt that exploits self-supervised learning in the differentiable architecture search domain~\cite{liu_2020_unnas}. While interesting, their approach is still within the scope of classification labels, \ie, to replace 
the human-annotated labels
with generated ones by tasks like
the Rotation~\cite{gidaris_2018_rotation}, Colorization~\cite{zhang_2016_col}, and solving Jigsaw puzzles~\cite{noroozi_2016_jigsaw} (middle block of \cref{fig:fig1}). 
We conjecture this is why their approach cannot even surpass the original DARTS.
By contrast, our method is fundamentally different from theirs in two-fold: 
i) we go beyond the traditional classification labels to let the network recovers masked patches, forcing the shared network to learn occlusion invariant information;
ii) we did not drop to use human-annotated labels to provide sufficient guidance on downstream tasks.  

We empirically demonstrate the effectiveness of our approach on popular differentiable architecture search benchmarks. 
On the DARTS search space, our searched final architecture has top-1 accuracy of 76.5\% on ImageNet and constitutes state-of-the-art performance when compared to previous DARTS methods.
It also outperforms all DARTS variants on CIFAR-10 and CIFAR-100 with average top-1 accuracy of 97.54\% and 83.81\%, respectively.
Moreover, our method enhances the generalization ability of the original DARTS and also maintains low search expenses.

In summary, our contributions are summarized as follow:
\begin{itemize}[leftmargin=8pt,itemsep=-1pt,topsep=0pt]
  \item We propose the MIM-DARTS that steps out of the scope of traditional differentiable architecture search, which only relies on the classification labels, and formulate a masked image modeling auxiliary loss to solve the collapse issue and enhance the performance.
  \item We empirically show that our framework achieves state-of-the-art performance on various settings while did not introduce any additional optimization tricks that are specifically designed for the original DARTS.
\end{itemize}
\vspace{-4pt}

\section{Related Works}
\label{sec:related}
\vspace{-4pt}
\mypara{Issues and solutions of DARTS.}
DARTS endures several optimization issues, of which Yu \etal questioned that the performance of the original DARTS cannot even surpass the random search~\cite{Yu2020Evaluating}.
Zela~\etal further identified a performance collapse that the obtained architecture is dominated by the parameter-free operations, \eg, \textit{skip} connections~\cite{Zela2020Understanding}.
To solve these issues, RobustDARTS~\cite{Zela2020Understanding} and SDARTS~\cite{chen_2020_sdarts} take the Hessian eigenvalue regarding $\alpha$ as an indicator for the occurrence of the collapse and adopt early-stop or perturbation to regularize the search process.
However, the quality of the indicator is still questioned.
P-DARTS~\cite{Chen_2019_pdarts} and DARTS+~\cite{liang2019darts+} propose to explicitly limit the number of the \textit{skip} connections, 
which may prevent DARTS from broadly exploring the search space.
FairDARTS~\cite{chu_2020_fairdarts} and DARTS-~\cite{chu2021dartsminus} avoids the unfair competition of \textit{skip} connections by an independent sigmoid function or additional \textit{skip} connections, which are hand-crafted.
Recently, Wang \etal discover that $\alpha$ cannot truthfully represent the operation strength,
and employ additional tuning to select architectures which increases the overheads and complexes the pipeline~\cite{wang_2021_rethinking}.
Even, $\beta$-DARTS~\cite{ye_2022_beta} proposes to directly regularize on the output of the softmax over $\alpha$ and achieves state-of-the-art.
A key fact is that DARTS suffers from insufficient training so as to favor architectures that converge fast~\cite{shu_2020_understanding}.
The aforementioned methods intentionally slow down the search pipeline by various regularizations.
Differently, our method proposes to inject more information into the optimization 
while avoids to complex the search pipeline and achieves state-of-the-art performance.

\mypara{Self-supervised learning in DARTS.}
There has been a preliminary attempt that enables the DARTS to learn semantic information by self-supervised learning~\cite{liu_2020_unnas}.
Despite their original goal to discuss whether labels are necessary for architecture search, 
we find that the semantic information, such as visuospatial representation, textures and colors, and geometric transformation learned by Rotation~\cite{gidaris_2018_rotation}, Colorization~\cite{zhang_2016_col} and solving Jigsaw puzzles~\cite{noroozi_2016_jigsaw},
is beneficial for solving collapse issue.
However, 
their optimization still maintains the classification, which may be insufficient to extract useful information, especially for architecture search.
Differently, 
our method is not limited to the traditional classification task and turns to learn occlusion invariant semantic information by MIM task,
which enables the architecture to learn more information than all previous DARTS variants.
Moreover, we do not drop to use the human-annotated labels to provide sufficient guidance on downstream tasks.

\mypara{Masked Image Modeling.}
MIM~\cite{Pathak_2016_Inpainting,He_2022_mae} task is a recent promising self-supervised learning method that reconstructs the masked patches of the image.
The pioneering work context encoder~\cite{Pathak_2016_Inpainting} masks out a square rectangular region and reconstructs the pixel colors via convolution architecture.
Current vision Transformer based methods exploit the representation learning ability of MIM by predicting the clustering of colors~\cite{chen_2020_igpt}, mean color~\cite{dosovitskiy2021vit}, the color of raw pixels\cite{He_2022_mae,Xie_2022_simmim} and patch tokens~\cite{bao2022beit}.
However, they focus on the representation learning and model weights pre-training,
while we are the first to bring it into the neural architecture search domain and testify its efficacy and key factors.
\vspace{-4pt}
\section{Method}
\label{sec:methods}
\vspace{-4pt}
In this section, we first 
revisit the formulation of
DARTS and showcase its performance collapse issue.
We then discover naively leveraging self-supervised learning tasks to replace human-annotated labels moderately addresses the collapse issue,
yet leads to sub-optimal performances. We propose a simple yet effective framework to incorporate 
Masked Image Modeling
task and go beyond traditional classification to kill two birds with one stone.
\begin{figure}[t]
    \centering
     \includegraphics[width=0.8\linewidth]{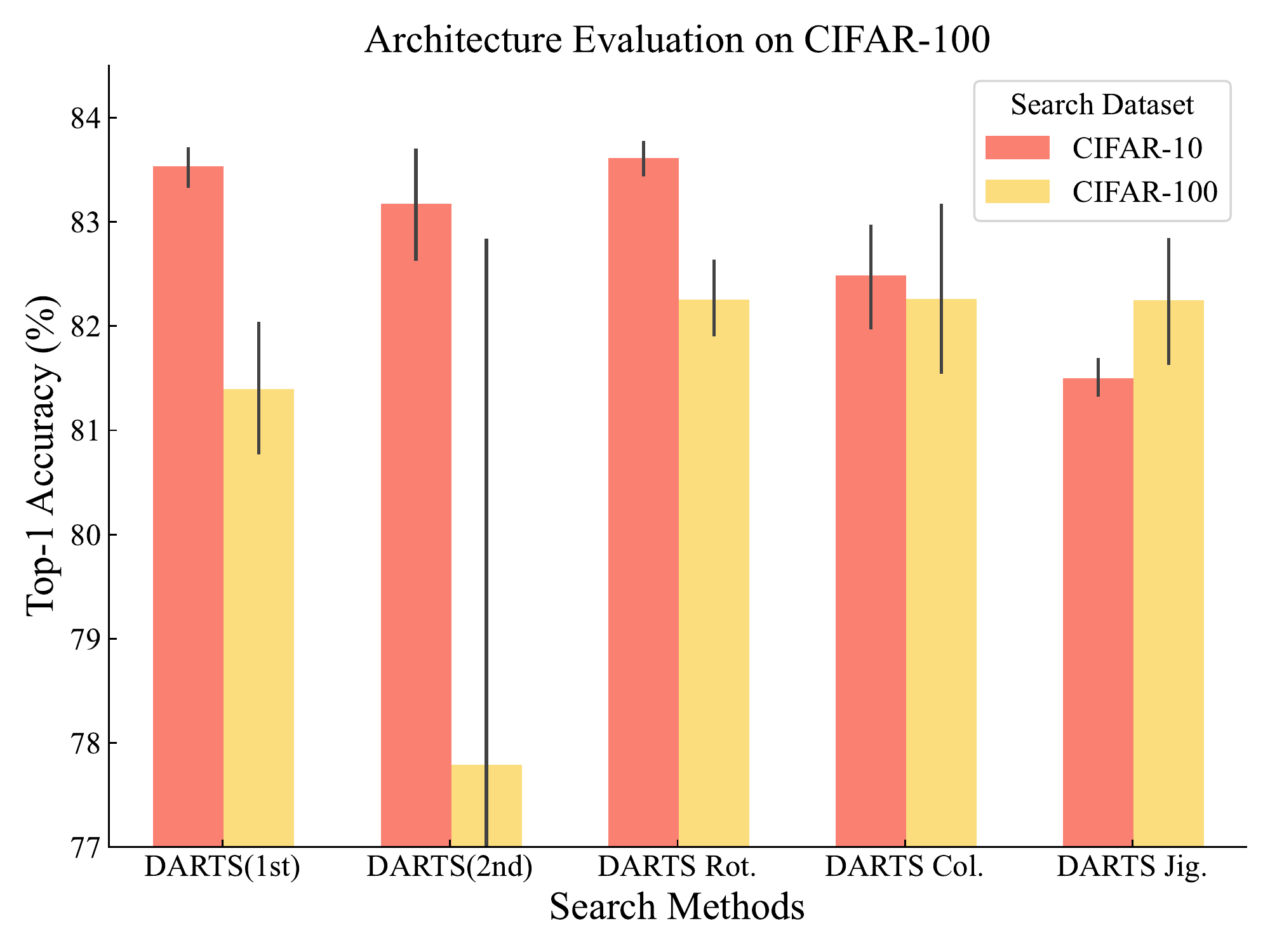}
     \vspace{-6pt}
     \caption{
    \textbf{Comparison of original DARTS and self-supervised DARTS. } 
    We run the DARTS with human-annotated classification labels and self-supervised generated tasks on CIFAR-10 and CIFAR-100, and evaluate the searched architecture on CIFAR-100. We notice that surprisingly, the searched models on the same dataset fall behind the one on CIFAR-10 by a margin of 3\% and 7\% for two DARTS settings. On the contrary, using self-supervised labels can significantly improve the CIFAR-100 searched architecture.
    }
     \label{fig:fig2}
     \vspace{-8pt}
  \end{figure}
  
\subsection{Preliminary}
\vspace{-4pt}
\label{sec:prel}
\mypara{Formulation of DARTS.}
We first give the formulation of the differentiable architecture search~\cite{liu2018darts}. Given a specific task, \eg, classification,
with a training set $\mathcal{D}_{train}$ 
and a validation set $\mathcal{D}_{val}$, DARTS searches for a computation cell as the building block of the final architecture.
Specifically, a cell is represented with a directed acyclic graph (DAG) consisting of $N$ nodes, where each node $x^{i}$ denotes a latent representation. 
Each directed edge ($i, j$) is associated with an operations, $o^{(i, j)} \in \mathcal{O}$, where $\mathcal{O}$ denotes a set of candidate operations.
DARTS conducts a continuous relaxation on learnable architecture parameters $\alpha$ to mix the outputs of the operations:
\vspace{-4pt}
\begin{equation}
  \bar{o}^{(i, j)}(x) = \sum_{k\in\mathcal{O}}\frac{\exp(\alpha_{k}^{(i, j)})}{\sum_{k'\in\mathcal{O}}\exp(\alpha_{k'}^{(i, j)})}o(x),
  \label{eq:original_darts}
  \vspace{-4pt}
\end{equation}
where $\bar{o}$ is the mixed output of a cell.
During the search phase, the cells are stacked to form the data encoder $E(\alpha, w)$, including the $\alpha$ and weights $w$ of the operations. 
We combine it with the task-specific head $H_{cls}$ and alternately optimize $\alpha$ and $w$ with loss $\mathcal{L}$:
\vspace{-4pt}
\begin{equation}
  \begin{split}
  & \min_{\alpha} \mathcal{L}_{val}(E(w^{\ast}(\alpha), \alpha), H_{cls}, \mathcal{D}_{val}) \\
  & \text{s.t. } w^{\ast}(\alpha) = \arg\min_{w}\mathcal{L}_{train}(E(w,\alpha), H_{cls}, \mathcal{D}_{train}),
  \end{split}
  \vspace{-4pt}
  \label{eq:alternate}
\end{equation}
where $w^{\ast}$ is approximated by one-step forward or current $w$, denoted as 2nd and 1st, separately.
\begin{figure}[t]
    \centering
     \includegraphics[width=1.0\linewidth]{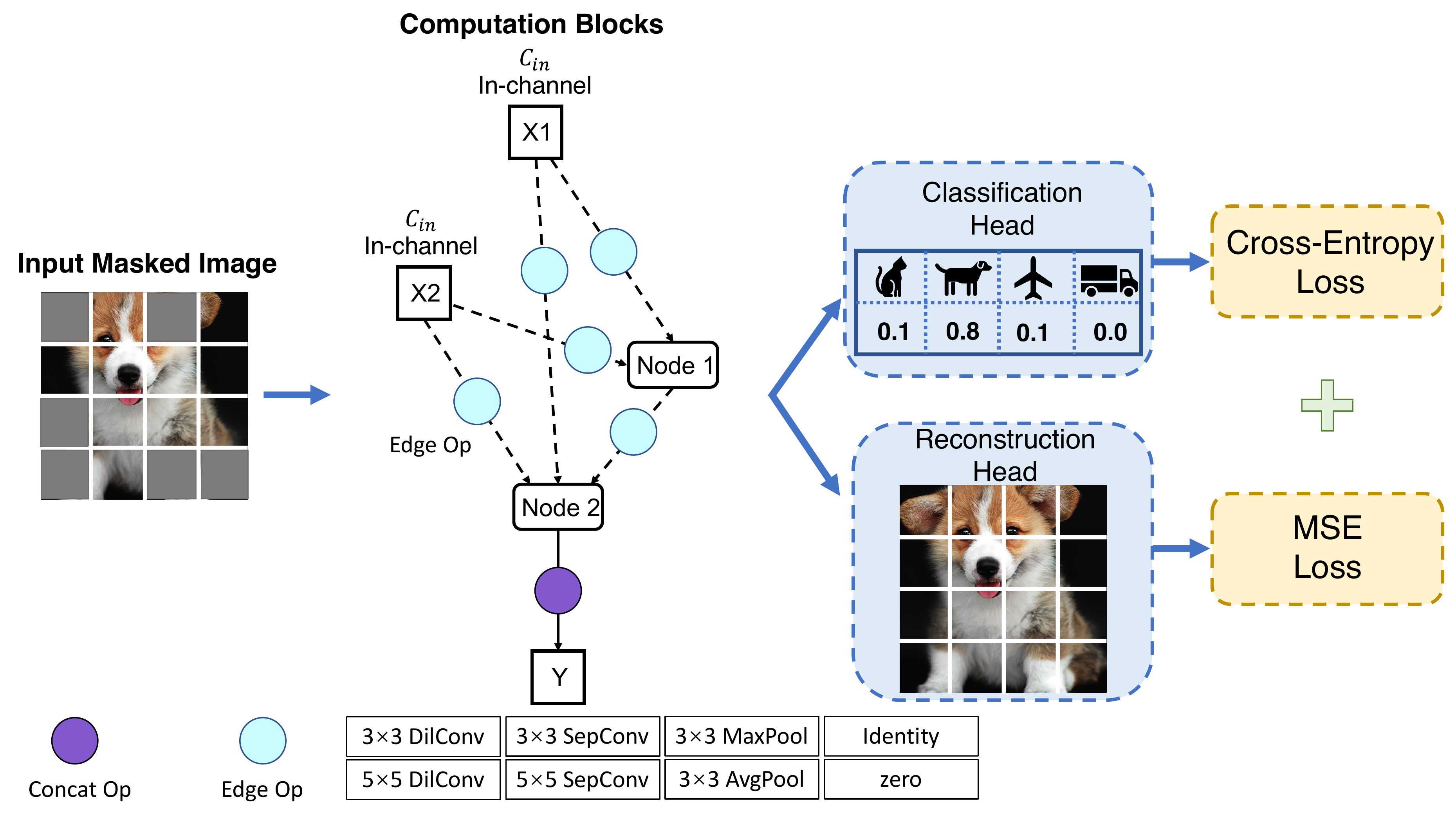}
     \vspace{-6pt}
     \caption{\textbf{The systematic illustration of the proposed method.} We perform jointly the classification and MIM tasks to inject class-specific and occlusion invariant semantic information into the architecture learning procedure. }
     \label{fig:fig3}
     \vspace{-8pt}
  \end{figure}
  
\mypara{Performance collapse issue.}
The original DARTS with the above formulation exhibits severe performance collapse issue: 
Zela \etal~\cite{Zela2020Understanding} found the architectures are dominated by \textit{skip} connections searched from 4 simplified CIFAR-10 DARTS search spaces.
As shown in \cref{fig:fig2}, we further identify the performance of the architectures searched from CIFAR-100 shows a large drop and fluctuation when performing one-step forward (2nd) approximation. 
Even for the more stable first-order approximation, more than 75\% of the operations collapse to $skip$ connections. 
The performance of the architectures searched on CIFAR-100 falls largely behind the ones searched on another dataset.

\mypara{Semantic information for DARTS.}
There has been a preliminary attempt that exploits self-supervised learning in DARTS domain~\cite{liu_2020_unnas}, trying to answer whether human annotations are necessary for architecture search task.
We find that the semantic information 
learnt by Rotaion~(Rot.)~\cite{gidaris_2018_rotation}, Colorization~(Col.)~\cite{zhang_2016_col} and solving the Jigsaw puzzles~(Jig.)~\cite{noroozi_2016_jigsaw},
is beneficial for DARTS to search stably and generalize well across the datasets.
As shown in the right part of \cref{fig:fig2}, when searching with self-supervised objectives, no performance collapse is observed and the performance of the architectures maintains basically the same level, despite the datasets searched from. 
This indicates that semantic information moderately helps to solve the collapse issue.
However, their performance can not even surpass the original DARTS, not to mention the recent state-of-the-art DARTS variants.
Meanwhile, for the Rotation task, the performance gap is still considerable, \ie, 82.24\%(C100)\textless83.60\%(C10).
We observe that a key similarity between self-supervised learning and all other DARTS variants is their optimization task is still within the classification scope: either human-annotated classification labels or self-supervised generated labels.
We conjecture that the single classification paradigm may be not adequate for architecture search to extract useful semantic information and leads to sub-optimal performance.

\subsection{MIM-DARTS}
\vspace{-4pt}
\label{sec:2tasks}
As Shu \etal~\cite{shu_2020_understanding} discover that insufficient training is the main cause of the collapse issue,
previous DARTS methods address it by intentionally slow-down the training process by various regularizations~\cite{chen_2020_sdarts,ye_2022_beta,liang2019darts+,Chen_2019_pdarts}.
Unfortunately, people focus solely on the classification domain to increase the learning ability which may be limited. To this end, we exploit an alternative approach to inject information other than classification labels while do not prolong the training process.
Specifically, we learn additionally the occlusion invariant information via a Masked Image Modeling~(MIM) task.
In principle, we formulate loss to reconstruct the random patches of the images to inject the occlusion invariant information.
Moreover, we do not drop the human-annotated labels to offer more guidance to the architecture.
The systematic illustration of the proposed method is shown in \cref{fig:fig3}.
The key components are discussed below.

\mypara{Masking.}
We change the clean image into a masked image to perform two tasks.
Specifically, the image $\bm{x} \in \mathbb{R}^{H \times W \times C}$ is divided into regular non-overlapping patches $\bm{x}_{p} \in \mathbb{R}^{N \times (P^{2}\cdot C)}$ where $P$ is the patch resolution and $N = HW/P^{2}$
, following ViT~\cite{dosovitskiy2021vit}.
We then uniformly sample a fixed ratio of patches without replacement, and "dropped out" them, \ie, set to zero, the same with context encoders\cite{Pathak_2016_Inpainting}.
The masking implementation follows~\cite{He_2022_mae}:
\begin{equation}
  \bm{x}_{mask} = (1-\bm{m})\odot \bm{x}_p,
  \vspace{-4pt}
  \label{eq:mask}
\end{equation}
where $\bm{x}_{mask} \in \mathbb{R}^{N \times (P^{2}\cdot C)}$ is the sequence of the masked patches; $\bm{m}$ is binary mask with the same shape of $\bm{x}_{p}$ in which value 1 indicates the masked pixel; $\odot$ is the element-wise product operation.
Different from \cite{He_2022_mae}, we reshape $\bm{x}_{mask}$ into a masked images as input $\bm{x}_{input} \in \mathbb{R}^{H \times W \times C}$.

\mypara{Encoder.}
We stack the computation cells into the data encoder $E$, serving as the masked image processor and searching for an optimal architecture during the optimization.
Besides the normal cell, there are reduction cells to reduce the spatial resolution of the latent feature whose candidate operations are of stride two.
\vspace{-4pt}
\begin{equation}
  \bm{x}_{inter} = E(\bm{x}_{input}),
  \vspace{-4pt}
  \label{eq:encoder}
\end{equation}
where $\bm{x}_{inter} \in \mathbb{R}^{H' \times W' \times C'}$ is the intermediate feature.

\mypara{Reconstruction decoder.}
The decoder $H_{\text{MIM}}$ takes the $\bm{x}_{inter}$ as input and reconstructs the masked patches according to the information from nearby visible patches.
Following \cite{Pathak_2016_Inpainting}, we first propagate the information across spatial by a 3$\times$3 convolution, and channels by a 1$\times$1 convolution.
Then we apply 2 transposed convolutions to gradually recover the spatial resolution to the size of the input image, before each we insert a learned filter followed by batch normalization and ReLU activation.
Finally, a 3$\times$3 convolution outputs a feature encoding the 3 channels of color values, and an element-wise HardTanh function is adopted to clip values to reasonable intervals.
\vspace{-4pt}
\begin{equation}
  \bm{x}_{rec} = H_{\text{MIM}}(\bm{x}_{inter}),
  \vspace{-4pt}
  \label{eq:decoder}
\end{equation}
where $\bm{x}_{rec} \in \mathbb{R}^{H \times W \times C}$ is the reconstructed image.

\mypara{Reconstruction target.}
We compute mean squared error~(MSE) over the reconstructed and original images. The computation limits to only the masked patches:
\vspace{-4pt}
\begin{equation}
  \mathcal{L}_{\text{MSE}} =  {\lVert \bm{m}\odot(\bm{x}_{rec} - \bm{x})\rVert}_2.
  \vspace{-4pt}
  \label{eq:mse}
\end{equation}

\mypara{Learning classification and MIM jointly.}
Besides the MIM task, we remain to perform classification to learn to extract the class-specific information at the same time:
$\bm{p} = H_{cls}(\bm{x}_{inter})$ and adopt cross-entropy loss $\mathcal{L}_{cls}$ to supervise the training. Thus, we can learn two tasks jointly and without losing simplicity and flexibility:
\vspace{-4pt}
\begin{equation}
  \mathcal{L} =  \mathcal{L}_{cls} + \lambda\mathcal{L}_{\text{MSE}},
  \vspace{-4pt}
  \label{eq:joint}
\end{equation}
where $\lambda$ is a trade-off parameter. Here we introduce an easy implementation to ease the weighting scheme. We view two types of tasks share the same significance.
And we align two losses by setting $\lambda = ({\mathcal{L}_{cls}}/{\mathcal{L}_{\text{MSE}}})$, where the computation of $\lambda$ does not involve in the gradients update. 
\vspace{-4pt}
\section{Experiments}
\vspace{-4pt}
\label{sec:exp}
\begin{table}
  \centering
  \caption{\textbf{Comparison results on CIFAR.} The results in the top block are obtained by training the best searched architecture; 
  the middle block is the average results for multiple runs of search.
  C10 and C100 denote the architectures are searched from CIFAR-10 and CIFAR-100 respectively.
All of our results are averaged by best searched architectures of four independent runs. 
  }
  \vspace{-6pt}
  \resizebox{\linewidth}{!}{
  \begin{tabular}{lccccc}
    \toprule
    \multirow{2}{*}{Method}   \hspace{-10pt}& Search Cost   & \multicolumn{2}{c}{CIFAR-10}    & \multicolumn{2}{c}{CIFAR-100}  \\ 
    \cmidrule(r){3-6}
                              & (GPU-Days)    & Params(M)     & Acc(\%)         & Params(M)     & Acc(\%)        \\
    \midrule
    NASNet-A~\cite{Zoph_2018_nasnet}              & 2000          &     3.3       & 97.35           & 3.3           & 83.18             \\
    DARTS(1st)~\cite{liu2018darts}             & 0.2           &     3.4       & 97.00$\pm$0.14  & 3.4           & 82.46             \\
    DARTS(2nd)~\cite{liu2018darts}             & 0.9           &     3.3       & 97.24$\pm$0.09  &     -         & -                 \\
    SNAS~\cite{xie2018snas}                   & 1.5           &     2.8       & 97.15$\pm$0.02  & 2.8           & 82.45             \\
    GDAS~\cite{Dong_2019_gdas}                   & 0.2           &     3.4       & 97.07           & 3.4           & 81.62             \\
    P-DARTS(C100)~\cite{Chen_2019_pdarts}           & 0.3           &      -        & 97.38           & 3.6           & 84.08             \\
    P-DARTS(C10)~\cite{Chen_2019_pdarts}           & 0.3           &     3.4       & 97.50           &               & 82.80             \\
    PC-DARTS\cite{Xu2020PC-DARTS}               & 0.1           &     3.6       & 97.43$\pm$0.07  & 3.6           & 83.10             \\
    CyDAS~\cite{Yu_2022_cdarts}                  & 0.3           &     3.6       & 97.60           & -             & -                 \\  
    \midrule
    P-DARTS~\cite{Chen_2019_pdarts}                & 0.3           & 3.3$\pm$0.21  & 97.19$\pm$0.14  & -             & -                 \\
    R-DARTS(L2)~\cite{Zela2020Understanding}            & 1.6           & -             & 97.05$\pm$0.21  & -             & 81.99$\pm$0.26    \\
    SDARTS-ADV~\cite{chen_2020_sdarts}             & 1.3           & 3.3           & 97.39$\pm$0.02  & -             & -                 \\
    DARTS+PT~\cite{wang_2021_rethinking}               & 0.8           & 3.0           & 97.39$\pm$0.08  & -             & -                 \\
    DARTS-~\cite{chu2021dartsminus}                 & 0.4           & 3.5$\pm$0.13  & 97.41$\pm$0.08  & 3.4           & 82.49$\pm$0.25    \\
    $\beta$-DARTS(C100)~\cite{ye_2022_beta}    & 0.4           & 3.78$\pm$0.08 & 97.49$\pm$0.07  & 3.83$\pm$0.08 & 83.48$\pm$0.03    \\
    $\beta$-DARTS(C10)~\cite{ye_2022_beta}     & 0.4           & 3.75$\pm$0.15 & 97.47$\pm$0.08  & 3.80$\pm$0.15 & 83.76$\pm$0.22    \\
    DOTS~\cite{Gu_2021_dots}                   & 0.3           & 3.5           & 97.51$\pm$0.06  & 4.1           & 83.52$\pm$0.13    \\
    CyDAS~\cite{Yu_2022_cdarts}                  & 0.3           & 3.9$\pm$0.08  & 97.52$\pm$0.04  & -             & 84.31              \\    
    \midrule
    ours(C100)             & 0.2           & 3.61$\pm$0.24 & 97.43$\pm$0.11  & 3.66$\pm$0.24 & \textbf{83.71$\pm$0.66} \\
    ours(C100 best)        & 0.2           & 3.73          & 97.59           & 3.78          & \textbf{84.45}                  \\  
    ours(C10)              & 0.2           & 4.05$\pm$0.23 & \textbf{97.54$\pm$0.15} & 4.10$\pm$0.23 & \textbf{83.81$\pm$0.44}  \\
    ours(C10 best)         & 0.2           & 4.10          & \textbf{97.71}            & 4.05          & 84.23              \\  
    \bottomrule
  \end{tabular}
  }
\label{tab:results_c10andc100}
\vspace{-6pt}
\end{table}
In this section, we conduct extensive experiments to verify the state-of-the-art performance and generalization ability of our simple method.
We further give detailed ablations of the key components and design insights.
Following the routine~\cite{liu2018darts}, each experiment consists of: 
i) search for an optimal cell by minimizing the validation loss;
ii) stack the cells into a full architecture and train it from the scratch for evaluation.
The experiments are performed on three datasets, CIFAR-10, CIFAR-100, and ImageNet~\cite{Deng_2009_ImageNet}, and two search spaces, DARTS~\cite{liu2018darts} and NAS-Bench-201~\cite{Dong2020nas201}.
Codes will be released at \url{https://github.com/AlbertiPot/MIM-DARTS}.
\begin{figure}
  \centering
  \begin{subfigure}{0.8\linewidth}
    \centering
    \includegraphics[width=0.8\linewidth]{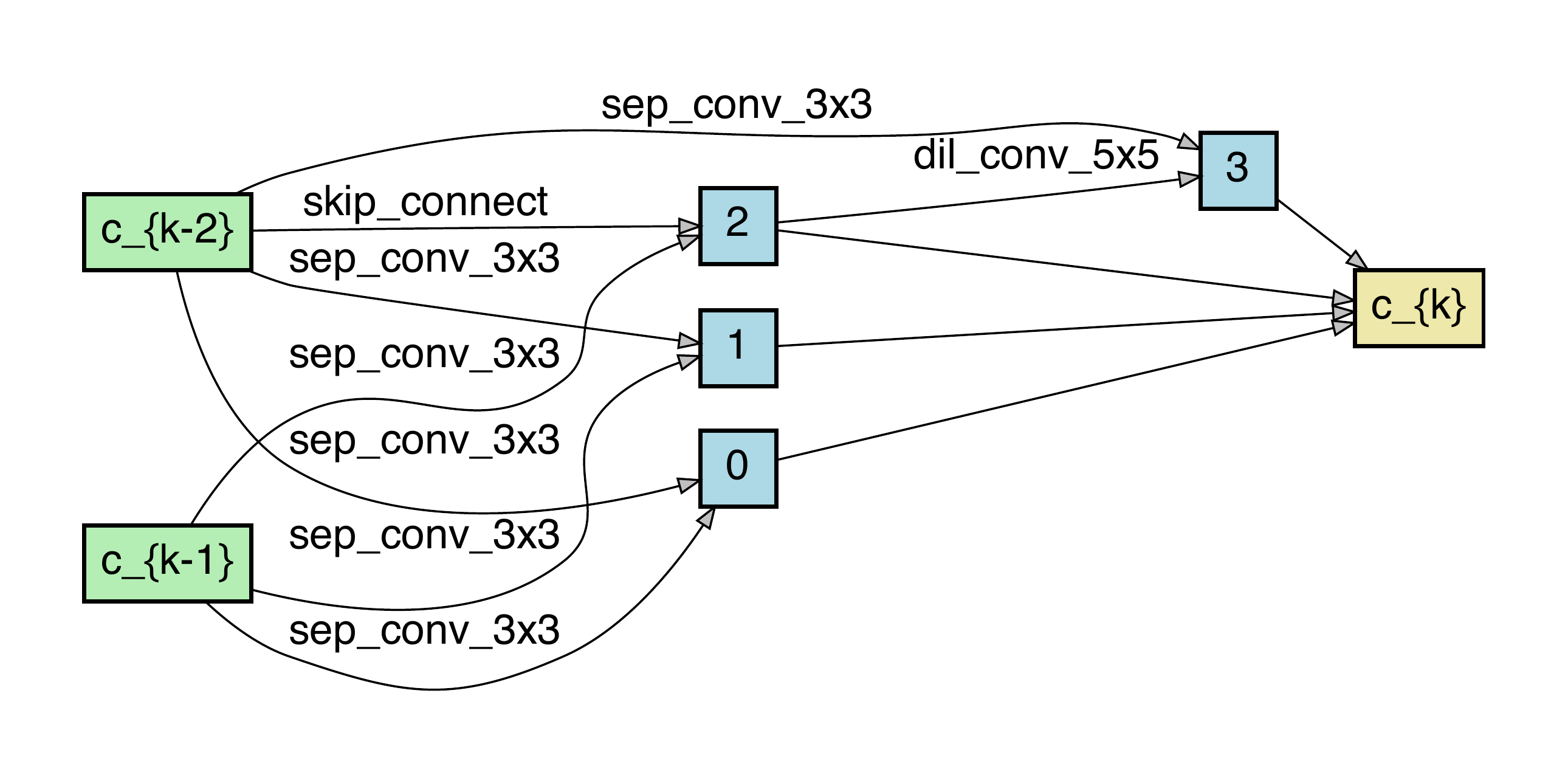}
    \vspace{-6pt}
    \caption{Normal Cell}
  \end{subfigure}

  \begin{subfigure}{0.8\linewidth}
    \centering
    \includegraphics[width=0.8\linewidth]{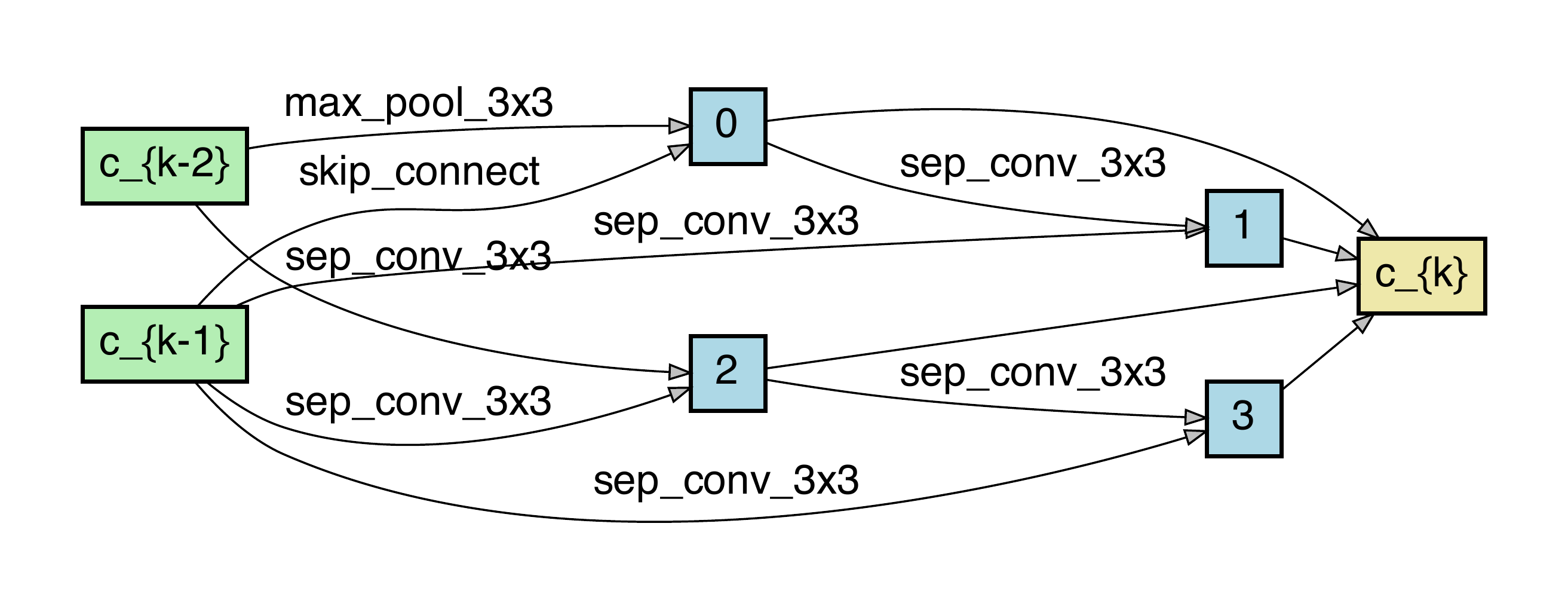}
    \vspace{-6pt}
    \caption{Reduction Cell}
  \end{subfigure}
  \vspace{-6pt}
  \caption{\textbf{Computation cells discovered on CIFAR-10}. 
  }
  \vspace{-6pt}
  \label{fig:arch_vis_search_c10}
\end{figure}
\subsection{Comparing with state-of-the-art methods}
\vspace{-4pt}
\label{sec:sota}
\mypara{Settings.}
In this section, we compare with previous state-of-art methods to demonstrate the efficacy of the proposed method.
We search architectures and evaluate on:
i) DARTS search space with CIFAR-10 and CIFAR-100;
2) NAS-Bench-201 search space with CIFAR-10.
DARTS 
is prevailing and challenging for evaluating the NAS algorithm of which each cell contains 4 intermediate nodes and 14 edges, each edge is connected with 8 candidate operations.
NAS-Bench-201~\cite{Dong2020nas201} is another prevailing cell-based search space.
Differently, each cell has 4 intermediate nodes and each edge is associated with 5 operations.
The search space is relatively small, containing 15,625 architectures with ground-truth performance.
The experiments on DARTS and NAS-Bench-201 follow~\cite{liu2018darts} and \cite{ye_2022_beta}.

\begin{table}
  \begin{center}
  \caption{\textbf{Comparison results on NAS-Bench-201~\cite{Dong2020nas201}.} Optimal denotes the best performing architecture in the search space.
          $\dagger$ denotes our reproduced results.
          All of our results are averaged by best searched architectures of four independent runs. 
          }
  \vspace{-6pt}
  \resizebox{0.8\linewidth}{!}{
    \begin{tabular}{lccc}
      
      \toprule
      \multirow{2}{*}{Methods}                        & Search Cost   & \multicolumn{2}{c}{CIFAR-10}    \\
      \cmidrule(r){3-4}
                                                      & GPU-Hours     & valid(\%)           & test(\%)            \\
      \midrule
      DARTS(1st)~\cite{liu2018darts}                  & 3.2           & 39.77$\pm$0.00          & 54.30$\pm$0.00          \\
      DARTS(2nd)~\cite{liu2018darts}                  & 10.2          & 39.77$\pm$0.00          & 54.30$\pm$0.00          \\
      GDAS~\cite{Dong_2019_gdas}                      & 8.7           & 89.89$\pm$0.08          & 93.61$\pm$0.09          \\
      SNAS~\cite{xie2018snas}                         & -             & 90.10$\pm$1.04          & 92.77$\pm$0.83          \\
      DSNAS~\cite{Hu_2020_dsnas}                      & -             & 89.66$\pm$0.29          & 93.08$\pm$0.13          \\
      PC-DARTS~\cite{Xu2020PC-DARTS}                  & -             & 89.96$\pm$0.15          & 93.41$\pm$0.30          \\
      iDARTS~\cite{zhang21s_idarts}                   & -             & 89.86$\pm$0.60          & 93.58$\pm$0.32          \\
      DARTS-~\cite{chu2021dartsminus}                 & 3.2           & 91.03$\pm$0.44          & 93.80$\pm$0.40          \\
      CyDAS~\cite{Yu_2022_cdarts}                     & -             & 91.12$\pm$0.44          & 94.02$\pm$0.31          \\
      $\beta$-DARTS$\dagger$~\cite{ye_2022_beta}      & 2.8           & 91.47$\pm$0.17          & 94.23$\pm$0.27          \\
      ours                                            & 2.9           & \textbf{91.49$\pm$0.12} & \textbf{94.30$\pm$0.13} \\
      \midrule
      optimal                                 & -       & 91.61               & 94.37               \\\hline 
    \end{tabular}
      }
    \label{tab:nas201}
  \end{center}
  \vspace{-6pt}
\end{table}
\begin{figure}
  \centering
  \begin{subfigure}{0.8\linewidth}
    \centering
    \includegraphics[width=0.9\linewidth]{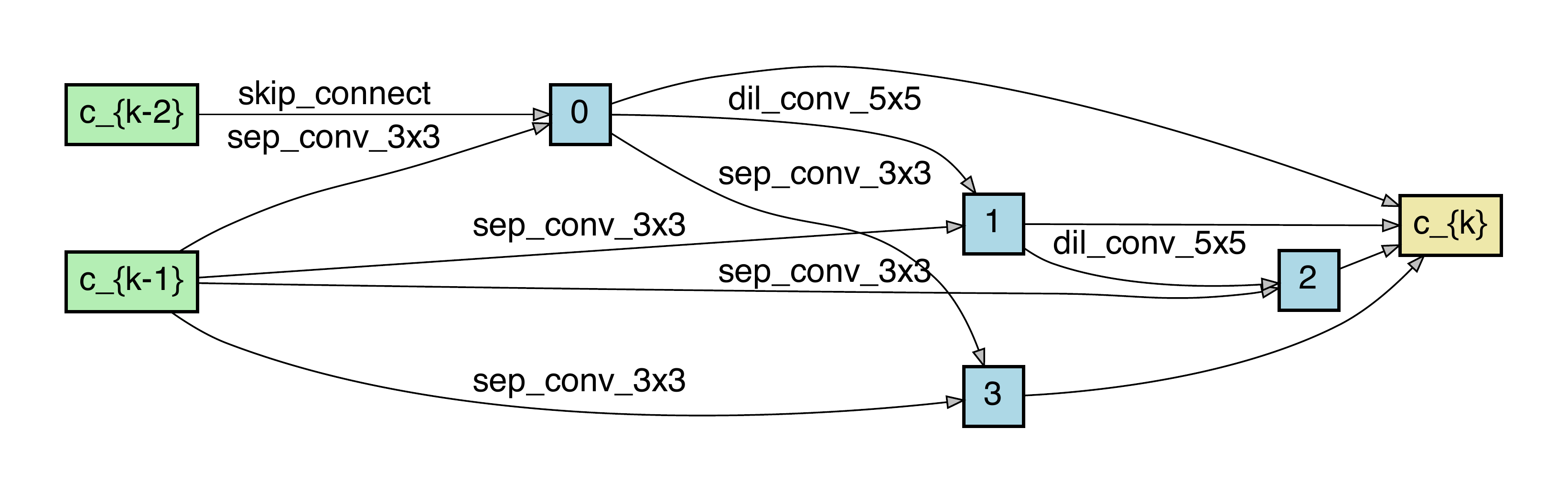}
    \vspace{-6pt}
    \caption{Normal Cell}
  \end{subfigure}

  \begin{subfigure}{0.8\linewidth}
    \centering
    \includegraphics[width=0.9\linewidth]{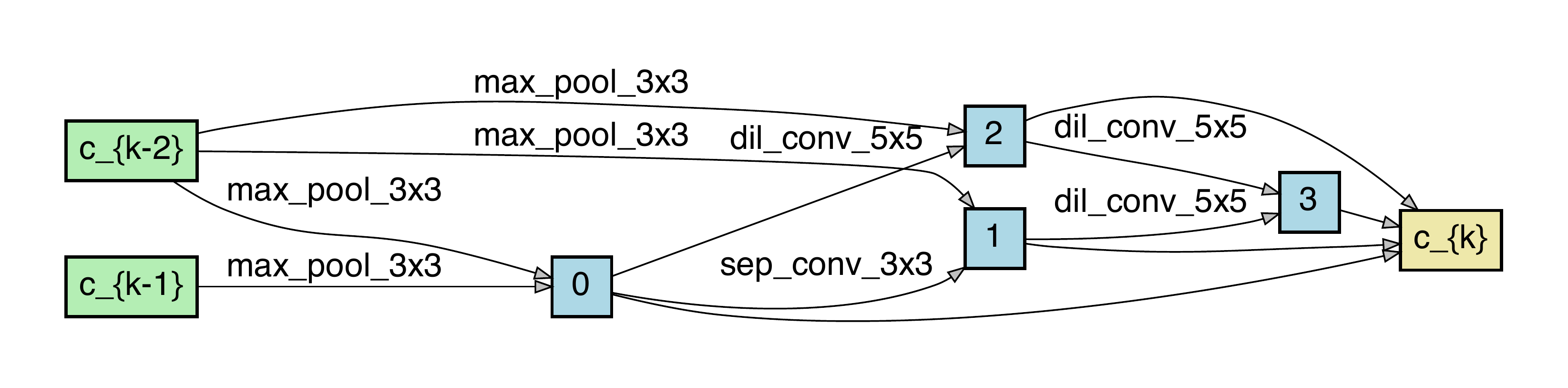}
    \vspace{-6pt}
    \caption{Reduction Cell}
  \end{subfigure}
  \vspace{-6pt}
  \caption{\textbf{Computation cells discovered on CIFAR-100}}
  \vspace{-6pt}
  \label{fig:arch_vis_search_c100}
\end{figure}
\mypara{Results on DARTS.}
The comparison results of the obtained architectures are shown in \cref{tab:results_c10andc100}.
For all 4 runs on CIFAR-10 and CIFAR-100 with same experiments settings in \cref{fig:fig2}, the performance collapse is well addressed:
the average performance of the architectures searched from CIFAR-100 (83.71\%) does not drop largely behind the ones searched from CIFAR-10 (83.81\%).
The searched architectures are not dominated by the $skip$ connections as shown \cref{fig:arch_vis_search_c10} and \cref{fig:arch_vis_search_c100}.
Moreover, the average accuracy of 4 runs (83.71\%) and the best accuracy (84.45\%) reach the highest among the CIFAR-100 counterparts with relative little parameters (average 3.66 million parameters)
We further achieve state-of-the-art average performance on CIFAR-10 and CIFAR-100, namely 97.54\% and 83.81\%. 
This demonstrates the stability of our 
framework
by adopting two distinct types of tasks which is superior than performing the traditional single classification task.
The best architecture from 4 runs achieves the highest performance among all previous DARTS variants, namely 97.71\% and 84.45\% on CIFAR-10 and CIFAR-100 respectively,
indicating the superior architecture search ability of the proposed method.

\mypara{Results on NAS-Bench-201.}
We further evaluate our method on NAS-Bench-201 search space by conducting 4 independent searches and evaluating the obtained architectures on CIFAR-10.
The comparison results are shown in \cref{tab:nas201}.
We achieve the state-of-the-art average validation and test accuracy, namely 91.49\% and 94.30\%.
The average accuracy is almost close to the optimal and shows a small fluctuation, indicating the ability of our method to search the superior architecture given a specific dataset.
\vspace{-4pt}
\begin{table}
    \centering
    \caption{\textbf{Comparison of generalization ability on CIFAR}. 
    \textbf{C10 to C100} denotes the architectures searched from CIFAR-10 are evaluated on CIFAR-100.
    average and best denote the average and best performance out of four independent runs.
    }
    \vspace{-6pt}
    \small
    \resizebox{0.8\linewidth}{!}{
    \begin{tabular}{lcccc}
    \toprule
    \multirow{2}{*}{Methods} & \multicolumn{2}{c}{C10 to C100} & \multicolumn{2}{c}{C100 to C10} \\
    \cmidrule(r){2-5}
                                        & average               & best          & average       & best      \\
    \midrule
    P-DARTS~\cite{Chen_2019_pdarts}     &         -             &    82.80      &         -                         &     97.38         \\
    DARTS+~\cite{liang2019darts+}       &         -             &    83.72      &         -                         &     97.54         \\
    $\beta$-DARTS~\cite{ye_2022_beta}   &        83.76$\pm$0.22 &     -         &        \textbf{97.49$\pm$0.07}             &    -              \\
    \midrule
    ours                                &       \textbf{83.81$\pm$0.44}  &      \textbf{84.23}    &         97.43$\pm$0.11            &     \textbf{97.59}         \\
    \bottomrule
    \end{tabular}
    }
    \vspace{-6pt}
    \label{tab:cifar_generalization}
\end{table}
\begin{table}
      \begin{center}
      \caption{\textbf{Comparison of generalization ability on NAS-Bench-201~\cite{Dong2020nas201}.} 
      All of our architectures are obtained by 4 independent runs of  search.
      Notice that the architectures are searched from CIFAR-10 and evaluated on CIFAR-100 and ImageNet-16-120.}
      \vspace{-6pt}
      \resizebox{\linewidth}{!}{
        \begin{tabular}{lcccc}
        
          \toprule
          \multirow{2}{*}{Methods}                              & \multicolumn{2}{c}{CIFAR-100}  & \multicolumn{2}{c}{ImageNet-16-120}\\
          \cmidrule(r){2-5}
                                                                           & valid(\%)           & test(\%)            & valid(\%)           & test(\%)         \\
          \midrule
          DARTS(1st)~\cite{liu2018darts}                                       & 15.03$\pm$0.00          & 15.61$\pm$0.00          & 16.43$\pm$0.00          & 16.32$\pm$0.00          \\
          DARTS(2nd)~\cite{liu2018darts}                                      & 15.03$\pm$0.00          & 15.61$\pm$0.00          & 16.43$\pm$0.00          & 16.32$\pm$0.00          \\
          GDAS~\cite{Dong_2019_gdas}                                          & 71.34$\pm$0.04          & 70.70$\pm$0.30          & 41.59$\pm$1.33          & 41.71$\pm$0.98          \\
          SNAS~\cite{xie2018snas}                                               & 69.69$\pm$2.39          & 69.34$\pm$1.98          & 42.84$\pm$1.79          & 43.16$\pm$2.64          \\
          DSNAS~\cite{Hu_2020_dsnas}                                             & 30.87$\pm$16.40         & 31.01$\pm$16.38         & 40.61$\pm$0.09          & 41.07$\pm$0.09          \\
          PC-DARTS~\cite{Xu2020PC-DARTS}                                       & 67.12$\pm$0.39          & 67.48$\pm$0.89          & 40.83$\pm$0.08          & 41.31$\pm$0.22          \\
          iDARTS~\cite{zhang21s_idarts}                                        & 70.57$\pm$0.24          & 70.83$\pm$0.48          & 40.38$\pm$0.59          & 40.89$\pm$0.68          \\
          DARTS-~\cite{chu2021dartsminus}                                     & 71.36$\pm$1.51          & 71.53$\pm$1.51          & 44.87$\pm$1.46          & 45.12$\pm$0.82          \\
          CyDAS~\cite{Yu_2022_cdarts}                                          & 72.12$\pm$1.23          & 71.92$\pm$1.30          & 45.09$\pm$0.61          & 45.51$\pm$0.72          \\
          $\beta$-DARTS$\dagger$~\cite{ye_2022_beta}                          & 73.02$\pm$0.95          & 73.10$\pm$0.82          & \textbf{46.22$\pm$0.31} & 45.92$\pm$0.85          \\
          ours                                                                & \textbf{73.21$\pm$0.56} & \textbf{73.16$\pm$0.70}  & 46.17$\pm$0.39          & \textbf{46.34$\pm$0.00} \\
          \midrule
          optimal                                                               & 73.49               & 73.51               & 46.77               & 47.31               \\ 
          \bottomrule
          \multicolumn{5}{l}{ \small
          Optimal denotes the best-performing architecture in the search space.} \\
           \multicolumn{5}{l}{ \small $\dagger$ denotes our reproduced results. }
        \end{tabular}      }
        \label{tab:nas201_transform}
      \end{center}
      \vspace{-4pt}
\end{table}

\subsection{Generalization ability of MIM-DARTS}
\vspace{-4pt}
\label{sec:generalization}
\mypara{Settings.}
We further extensively testify the generalization ability of the obtained architecture by our proposed method.
Specifically, for DARTS search space, we search on one dataset and evaluate the architectures on others, \eg, search from CIFAR-10 and evaluate on CIFAR-100 and ImageNet;
for NAS-Bench-201 search space, we transform the obtained architectures from CIFAR-10 to evaluate on CIFAR-100 and ImageNet-16-120.
The ImageNet experiments follow~\cite{Zhang_2021_rlnas} and the others maintain the same as \cref{sec:sota}.

\mypara{Results on CIFAR.}
The comparison of generalization ability on CIFAR is shown in \cref{tab:cifar_generalization}.
When evaluated on CIFAR-100, our method is competitive with previous SOTA~\cite{ye_2022_beta} on average 97.43\% and best one 97.59\%.
Further, the architectures searched from CIFAR-10 generalize well on CIFAR-100 by achieving state-of-the-art generalization performance on average 83.81\% and best one 84.23\%.
Thus, by learning the occlusion invariant information and class-specific feature, our framework enhances the generalization ability to a new high for original DARTS.
\begin{table}
  \centering
  \caption{\textbf{Comparison of generalization ability on ImageNet.} 
  The top three blocks denote the architectures are searched 1) from the ImageNet;
  2) with training samples in CIFAR-10 and parts of the ImageNet;
  3) from CIFAR-10 (C10) or CIFAR-100 (C100).}
  \vspace{-6pt}
  \resizebox{\linewidth}{!}{
  \begin{tabular}{lccccc}
    \toprule
    \multirow{2}{*}{Method}                               & Search Cost   & Params    & FLOPs   & \multicolumn{2}{c}{Test Err.} \\
    \cmidrule(r){5-6}
                                                          & (GPU-Days)    & (M)       & (M)     & Top1(\%)& Top5 (\%) \\
    \midrule
    MnasNet-92~\cite{Tan_2019_MnasNet}                               & 1667          & 4.4       & 388     & 74.8    & 92.0  \\
    FairDARTS~\cite{chu_2020_fairdarts}                              & 3             & 4.3       & 440     & 75.6    & 92.6   \\
    PC-DARTS~\cite{Xu2020PC-DARTS}                                        & 3.8           & 5.3       & 597     & 75.8    & 92.7    \\
    DOTS~\cite{Gu_2021_dots}                                       & 1.3           & 5.3       & 596     & 76.0    & 92.8   \\
    DARTS-~\cite{chu2021dartsminus}                                & 4.5           & 4.9       & 467     & 76.2    & 93.0    \\
    CyDAS\cite{Yu_2022_cdarts}                               & 1.7           & 6.1$\pm$0.2&701$\pm$32&76.3$\pm$0.3&92.9$\pm$0.2 \\
    \midrule
    AdaptNAS-S~\cite{Li_2020_AdaptNAS_nips}                                       & 1.8           & 5.0       & 552     & 74.7    & 92.2   \\
    AdaptNAS-C~\cite{Li_2020_AdaptNAS_nips}                                       & 2.0           & 5.3       & 583     & 75.8    & 92.6    \\ 
    \midrule
    AmoebaNet-C(C10)~\cite{Real_Aggarwal_Huang_Le_2019_AmoebaNet}                                      & 3150          & 6.4       & 570     & 75.7    & 92.4   \\
    SNAS(C10)~\cite{xie2018snas}                                             & 1.5           & 4.3       & 522     & 72.7    & 90.8   \\
    P-DARTS(C100)~\cite{Chen_2019_pdarts}                                         & 0.3           & 5.1       & 577     & 75.3    & 92.5   \\
    SDARTS-ADV(C10)~\cite{chen_2020_sdarts}                                       & 1.3           & 5.4       & 594     & 74.8    & 92.2   \\
    DOTS(C10)~\cite{Gu_2021_dots}                                             & 0.3           & 5.2       & 581     & 75.7    & 92.6   \\
    DARTS+PT(C10)~\cite{wang_2021_rethinking}                                         & 0.8           & 4.6       & -       & 74.5    & 92.0    \\
    $\beta$-DARTS(C100)~\cite{ye_2022_beta}                    & 0.4                           & 5.4       & 597     & 75.8    & 92.9     \\
    $\beta$-DARTS(C10)~\cite{ye_2022_beta}                    & 0.4                           & 5.5       & 609     & 76.1    & 93.0     \\
    \midrule
    ours(C100)                                            & 0.2           & 5.2       & 592     &  75.7   & 92.7 \\       
    ours(C10)                                             & 0.2           & 5.8       & 642     &  \textbf{76.5}   & 93.0     \\  
    \bottomrule
  \end{tabular}
  }
  \label{tab:results_img}
  \vspace{-6pt}
\end{table}

\mypara{Results on ImageNet.}
We further evaluate the generalization performance of the architectures searched from CIFAR-10 and CIFAR-100 on ImageNet.
As shown in \cref{tab:results_img}, with fewer search costs, our method shows very competitive generalization performance.
The one searched with only the training samples from the CIFAR-10 achieves new state-of-the-art top-1 accuracy with 76.5\% among the DARTS variants, 
overwhelming the counterparts that either introduce additional training samples from other domains~\cite{Li_2020_AdaptNAS_nips}, 
or have to be searched directly from the ImageNet which causes larger search expenses~(from 1.7 to 1667 GPU-Days compared to our 0.2 GPU-Days).

\mypara{Results on NAS-Bench-201.}
As shown in \cref{tab:nas201_transform}, our method shows well generalization ability on small search space.
Same with the observation on DARTS search space, the architectures searched from CIFAR-10 generalize well to CIFAR-100 by achieving the highest validation and test average accuracy, namely 73.21\% and 73.16\%.
On ImageNet-16-120, though it falls slightly (0.05\%) behind the recent SOTA~\cite{ye_2022_beta}, on the validation set, it still achieves the state-of-the-art on the test set, enhanced by 0.42\%.

\subsection{Comparing with self-supervised tasks}
\vspace{-4pt}
\begin{table*}
  \centering
  \caption{\textbf{Comparison results with classification-based self-supervised tasks.} 
  Cls. denotes performing the classification on CIFAR-10.
  Rot., Col., and Jig. denote performing the classification task with labels generated from Rotation~\cite{gidaris_2018_rotation}, Colorization~\cite{zhang_2016_col}, and solving the Jigsaw puzzles~\cite{noroozi_2016_jigsaw} tasks.
  All experiments are run 4 times independently on CIFAR-10 (C10) and CIFAR-100 (C100).
  }
  \vspace{-6pt}
  \small
  \resizebox{0.55\linewidth}{!}{
  \begin{tabular}{lccccc}
    \toprule
    \multirow{2}{*}{Methods}   & Search Cost   & \multicolumn{2}{c}{CIFAR-10}    & \multicolumn{2}{c}{CIFAR-100}  \\ 
    \cmidrule(r){3-6}
                              & (GPU-Days)    & Params(M)     & Acc(\%)         & Params(M)     & Acc(\%)        \\
    \midrule
    Cls.+ Rot.~(C100)   & 0.5           & 2.19$\pm$0.18 & 97.03$\pm$0.07  & 2.24$\pm$0.18 & 82.10$\pm$0.39    \\
    Cls.+ Col.~(C100)   & 0.2           & 2.09$\pm$0.29 & 97.05$\pm$0.22  & 2.14$\pm$0.29 & 81.73$\pm$0.74    \\
    Cls.+ Jig.~(C100)   & 0.2           & 4.31$\pm$0.29 & 97.02$\pm$0.04  & 4.36$\pm$0.29 & 82.40$\pm$0.39    \\
    ours~(C100)             & 0.2           & 3.61$\pm$0.24 & \textbf{97.43$\pm$0.11} & 3.66$\pm$0.24 & \textbf{83.71$\pm$0.66} \\
    \midrule
    Cls.+ Rot.~(C10)    & 0.5           & 3.44$\pm$0.54 & 97.35$\pm$0.12  & 3.49$\pm$0.54 & 83.57$\pm$0.64    \\
    Cls.+ Col.(C10)    & 0.2           & 2.37$\pm$0.33 & 97.19$\pm$0.25  & 2.42$\pm$0.33 & 82.12$\pm$0.81    \\
    Cls.+ Jig.~(C10)    &  0.2          & 4.48$\pm$0.25 & 96.91$\pm$0.43  & 4.53$\pm$0.25 & 81.72$\pm$1.11    \\
    ours~(C10)              & 0.2           & 4.05$\pm$0.23 & \textbf{97.54$\pm$0.15} & 4.10$\pm$0.23 & \textbf{83.81$\pm$0.44}  \\
    \bottomrule
\end{tabular}
}
\label{tab:unnas_with_multi}
\vspace{-6pt}
\end{table*}
\begin{figure*}
  \centering
  \begin{subfigure}{0.28\linewidth}
    \includegraphics[width=1\linewidth]{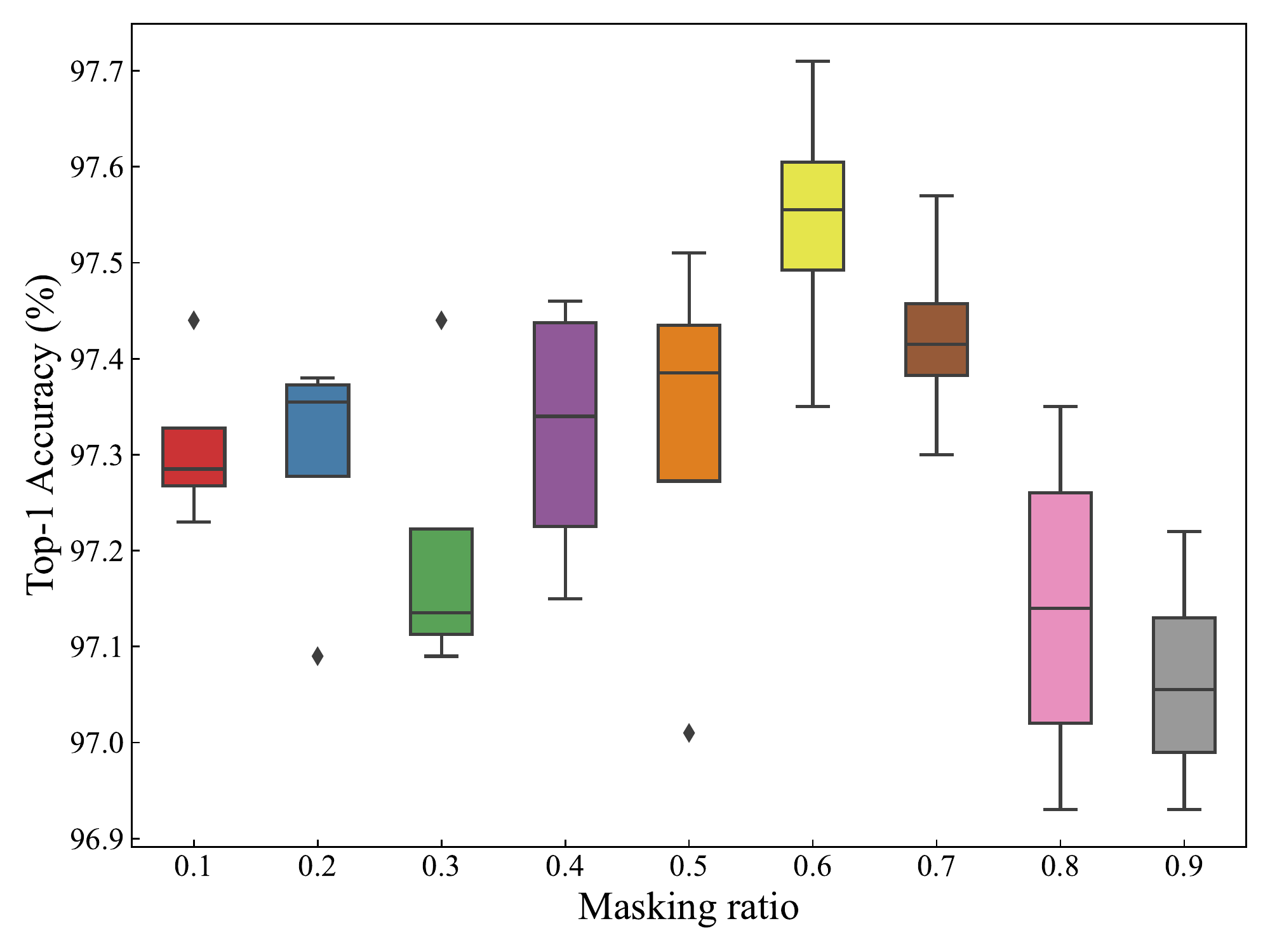}
    \caption{Patch size = 8.}
    \label{fig:ps8}
  \end{subfigure}
  \hspace{+8pt}
  \begin{subfigure}{0.28\linewidth}
    \includegraphics[width=1\linewidth]{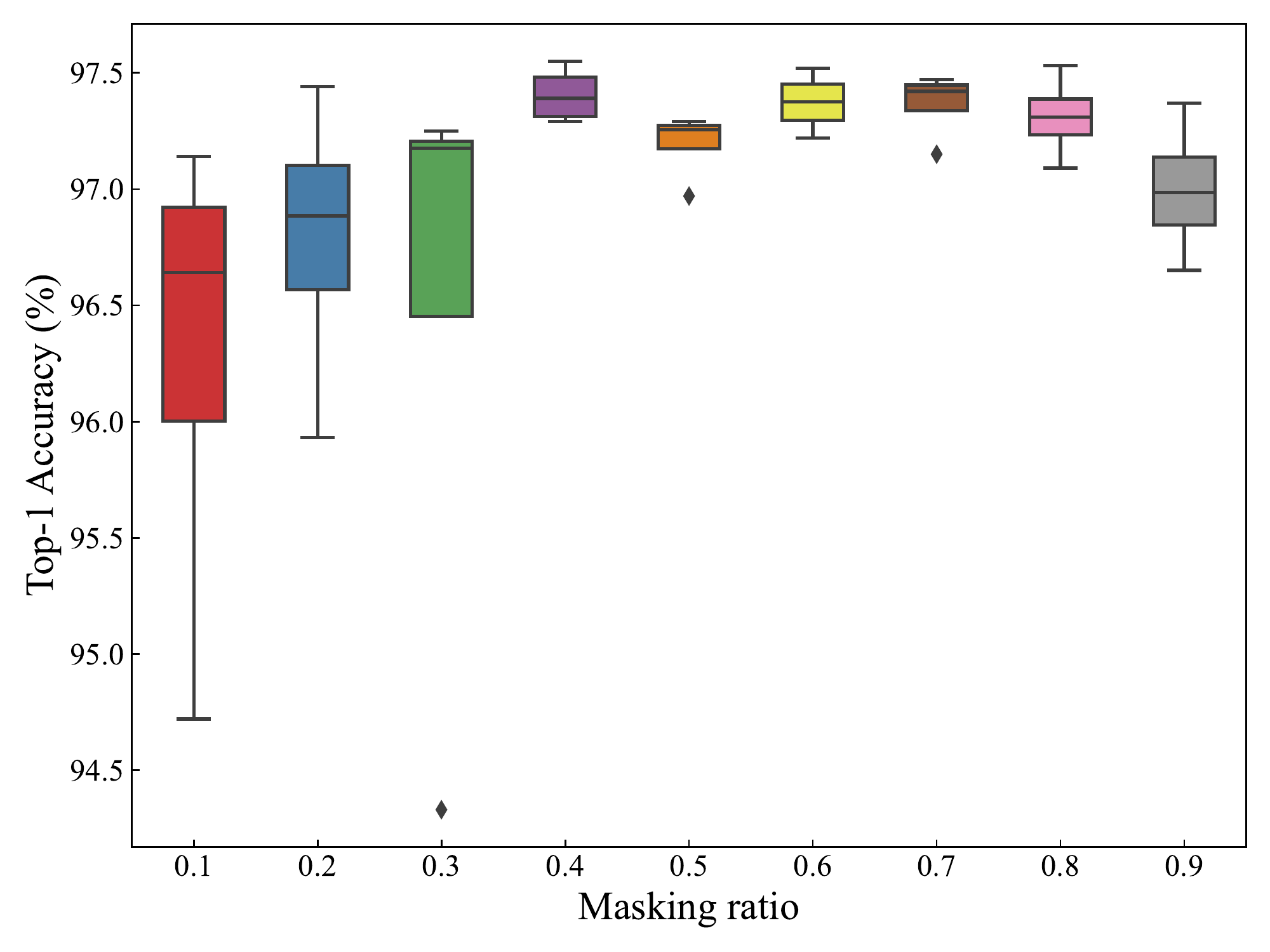}
    \caption{Patch size = 4.}
    \label{fig:ps4}
  \end{subfigure}
  \hspace{+8pt}
  \begin{subfigure}{0.28\linewidth}
    \includegraphics[width=1\linewidth]{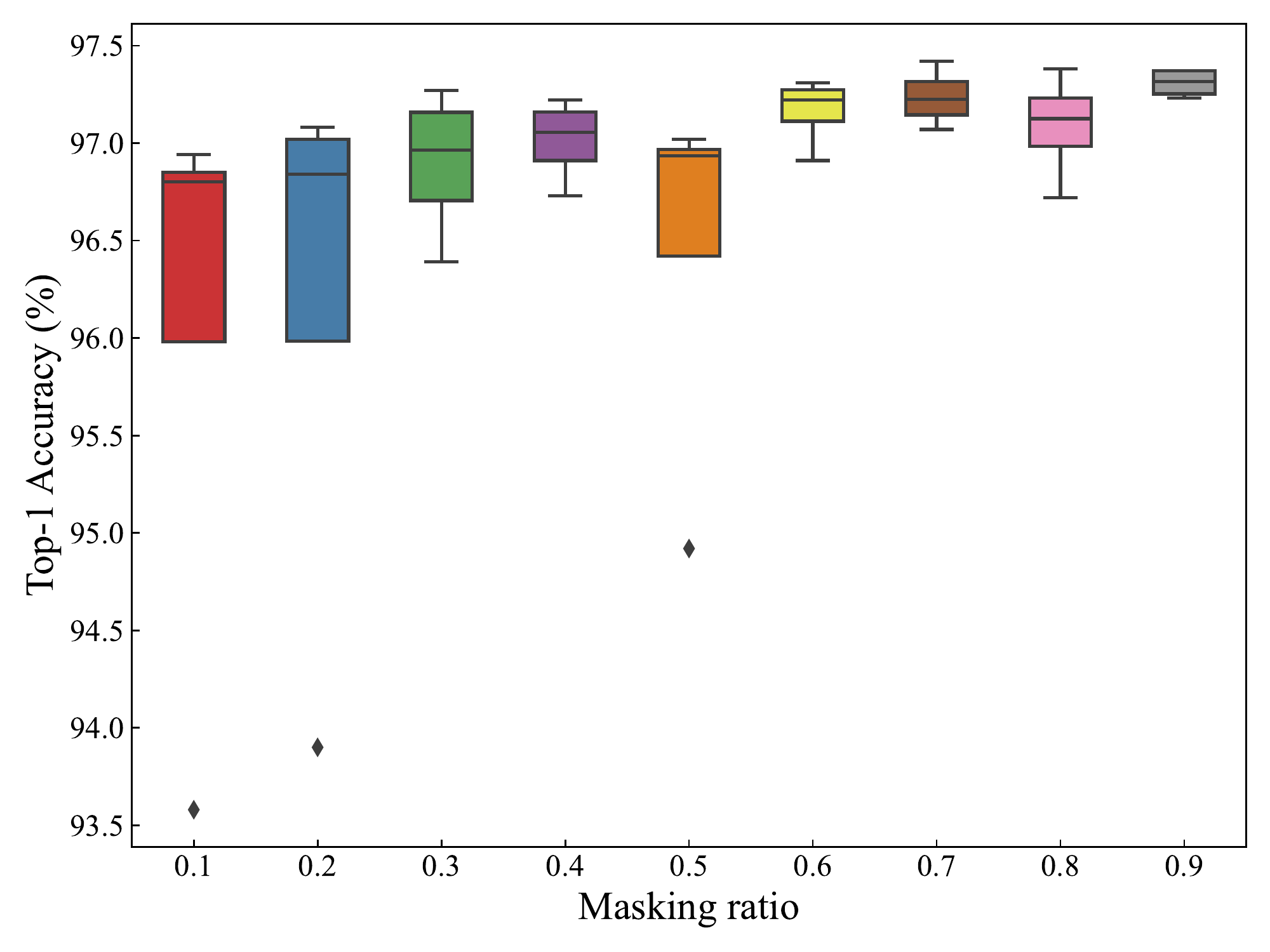}
    \caption{Patch size = 2.}
    \label{fig:ps2}
  \end{subfigure}
  \vspace{-6pt}
  \caption{\textbf{Ablations on different patch sizes and masking ratios.}
  The performance of the architectures saturates at middle masking ratio 0.6 and 0.4 when patch size is 8 and 4 respectively; 
  while at patch size 2, the performance continuously improves with the masking ratio.
  The best architectures are searched under patch size 8 and masking ratio 0.6.
  All experiments are run 4 times independently on CIFAR-10.}
  \label{fig:c10_patch_size}
  \vspace{-6pt}
\end{figure*}
In this section, we evaluate the capability of the MIM task to extract useful semantic information than previous self-supervised learning objectives.
As discussed in \cref{sec:prel}, the semantic information learned by Rot., Col., and Jig. cannot solve the collapse issue while enhance the performance at the same time.
We contribute this dilemma to their optimization is still performing the single task paradigm: \ie, classification of 4-way orientation, 313-way quantized color value, and 24-way permutations of the patches.
We verify this by performing the classification tasks of the original CIFAR and self-supervised objectives and comparing them with our proposed dual-task framework.
The settings follow \cite{liu_2020_unnas} and \cref{sec:sota}.

\mypara{Results.}
The comparison results of 4 independent runs are shown in \cref{tab:unnas_with_multi}.
We can see that, though the architectures are learned with two types of information (but with single learning task), the evaluation performance basically maintains the same performance as shown in \cref{fig:fig2}.
Besides, the top performance is the one searched under Rot. on CIFAR-10, yielding average accuracy of 97.35\% and 83.57\% on CIFAR-10 and CIFAR-100 respectively, but still falls behind ours and costs more search expenses.
This clearly indicates that, performing the original CIFAR classification task and self-supervised objectives such as Rot., Col., and Jig. simultaneously can not enhance the performance at all.
It further testifies our discussion that, no matter what objectives it searches for, the single knowledge handling method, \ie, classification, is not enough to inject the needed information into the insufficient architecture search procedure,
so as to enhance the performance.
However, by performing classification and MIM tasks simultaneously, we outperform all the classification-based self-supervised objectives and achieve state-of-the-art performance among the DARTS variants.
This demonstrates that:
1) the MIM task is more suitable for learning additional semantics information which helps to stabilize the search and enhance the performance in the meantime, compared to Rot., Col., and Jig.;
2) performing another type of knowledge-handling method besides the classification is a promising way to make up a deficiency that previous DARTS only learned abstract class-specific features.
We wish to open a door for discussing more practical knowledge-handling methods for architecture search to further boost the improvements.

\subsection{Ablation Studies}
\label{sec:ablation}
\begin{figure}[t]
  \centering
  \includegraphics[width=0.7\linewidth]{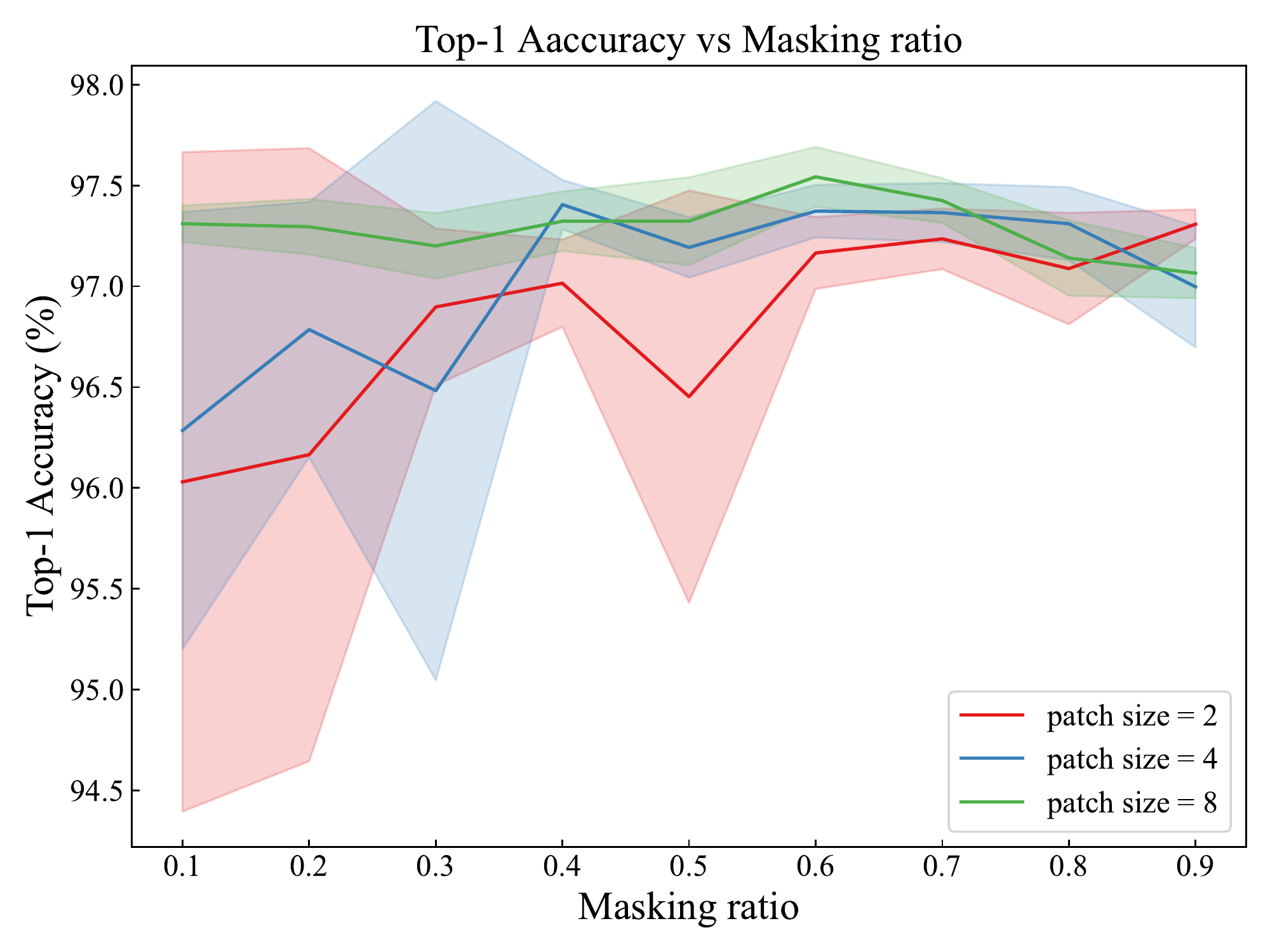}
    \vspace{-6pt}
  \caption{\textbf{CIFAR-10 top-1 classification accuracy under various masking ratios.}
  The best architectures searched from CIFAR-10 are at patch size 8 with masking ratio 0.6.
  At patch size 8, the performance of the architectures shows small variations, indicating a wide range of masking ratios that works well.
  }
  \label{fig:c10_patch_ratio}
  \vspace{-6pt}
\end{figure}

\begin{figure}[t]
  \centering
   \includegraphics[width=0.7\linewidth]{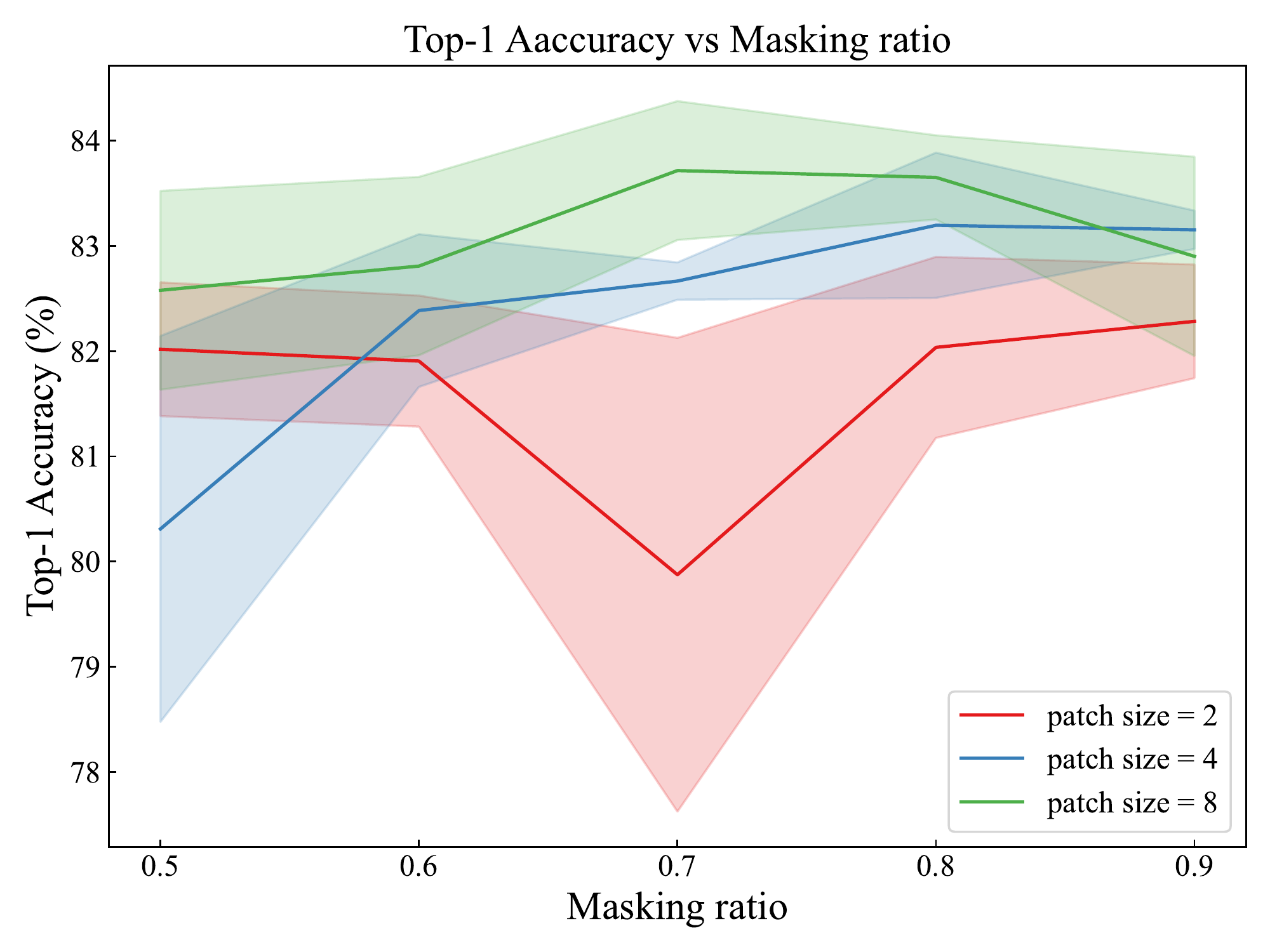}
    \vspace{-6pt}
   \caption{\textbf{CIFAR-100 top-1 classification accuracy under various masking ratios.}
   The best architectures searched from CIFAR-100 are at patch size 8 with masking ratio 0.6.}
   \label{fig:c100_patch_ratio}
   \vspace{-6pt}
\end{figure}

\vspace{-4pt}
\mypara{Patch size and masking ratio.}
We first explore the influence of various patch sizes and the masking ratios for CIFAR-10 and CIFAR-100 respectively.
In our implementation, the patch size is chosen from 2, 4, and 8, which are 256, 64, and 16 patches for the images of CIFAR in the shape of 32.
As shown in \cref{fig:c10_patch_size}, when at different patch sizes of 8, 4, and 2, the best architectures appear at masking ratios 0.6, 0.4, and 0.9, whose average top-1 accuracy is 97.54\%, 97.40\%, and 97.31\%, respectively.
We can see that when at lower patch size (2), the performance improves as the masking ratio grows, while at higher sizes (4 and 8), the performance will saturate at the middle masking ratio and start to drop at high.
This indicates that the small patch size is not sufficient for architectures to extract useful information for enhancing performance.
Moreover, by comparing different patch sizes and masking ratios for CIFAR-10 and CIFAR-100 in \cref{fig:c10_patch_ratio} and \cref{fig:c100_patch_ratio}, we observe that at patch size 8, 
the performance of various masking ratio is more stable and fluctuate smaller than 4, and 2, indicating a wide range of mask ratios that works well for large patch size.
The final optimal architectures are obtained at 1) patch size 8 and masking ratio 0.6 for CIFAR-10; 2) patch size 8 and masking ratio 0.7 for CIFAR-100, which are our baselines.

\begin{table}
  \centering
  \caption{\textbf{Ablations on key components.} \textbf{Cls.} denotes performing the classification task on CIFAR-10.
  Rec. denotes performing the clean image reconstruction task.
  mask denotes applying random masking following \cref{sec:2tasks}.}
  \vspace{-6pt}
  \small
  \resizebox{0.75\linewidth}{!}{
  \begin{tabular}{ccc|cc}
    \toprule
    Cls.        &   Rec.      &   mask      & Accuracy(\%)      & Params(M)             \\
    \midrule
    \checkmark  &            &              & 97.00$\pm$0.14    &   3.4                 \\
    \checkmark  &            & \checkmark   & 97.05$\pm$0.26    &   4.24$\pm$0.23       \\
                & \checkmark &              & 84.09$\pm$0.00    &   1.59$\pm$0.00       \\
                & \checkmark & \checkmark   & 97.18$\pm$0.12    &   2.46$\pm$0.35       \\

    \checkmark  & \checkmark & \checkmark   & \textbf{97.54$\pm$0.15}    &   4.05$\pm$0.23       \\
    \bottomrule
  \end{tabular}
  }
  \label{tab:lossandinput}
  \vspace{-6pt}
\end{table}

\mypara{In what situation MIM works for architecture search?}
We are the first to bring the MIM task to the architecture search domain.
It turns out that learning the class-specific and occlusion invariant semantic information simultaneously by adopting two types of knowledge acquisition methods works well for DARTS.
When only performing the MIM as shown in the fourth line of \cref{tab:lossandinput}, the average performance reaches to 97.18\% and outperforms the original DARTS performing the classification task.
This demonstrates that the MIM task is a promising way for extracting the required knowledge for the vision task.
However, it still falls behind recent DARTS variants and ours (97.54\%).
This echoes the proposal in \cref{sec:intro} that for architecture search, more information is required to search for a well-performed architecture.
So we choose to learn both the classification and MIM tasks simultaneously.

\mypara{Key components.}
We have further demonstrated the effectiveness of the critical components of our methods by gradually adding one part to the original DARTS as shown in \cref{tab:lossandinput}.
Notice that at the second line, performing the original CIFAR classification task on masked images is similar to applying Cutout regularization~\cite{Devries_2018_cutout}, except we mask out random patches rather than a square region.
However, it yields limited improvements (97.05\% vs 97.00\% of original DARTS).
When performing only the reconstruction task on the clean images, the obtained architectures are dominated by the \textit{skip} connections and it yields poor results (84.09\%).
This is natural because searching for an architecture that reconstructs an image is equal to learning an identity function.
Finally, when performing the classification and MIM tasks, the class-specific and occlusion invariant knowledge are injected into the insufficient architecture search procedure,
and our framework yields robust architectures that achieve superior performance among the DARTS variants.

\mypara{Search expenses}
The additional MIM branch does not introduce significant search expenses.
On DARTS search space, the proposed framework maintains the efficiency of the original DARTS, \ie, 0.2 GPU-Days as shown in \cref{tab:results_c10andc100}.
On the NAS-Bench-201 search space, it achieves almost the same low costs as \cite{ye_2022_beta}, \ie, 2.9 GPU-Hours in \cref{tab:nas201}.
Moreover, when learning with traditional classification together, our framework shows its superiority by achieving state-of-the-art results while maintaining low expenses in \cref{tab:unnas_with_multi}.
\vspace{-4pt}
\section{Conclusions}
\vspace{-4pt}
In this work, we take the first trail to inject more information into the insufficient learning procedure of the differentiable architecture search,
which is ignored by previous work.
By formulating an additional Masked Image Modeling task, we step out of the traditional single classification task while do not abandon the useful guidance from the downstream tasks.
Extensive experiments indicate our simple framework well addresses the notorious performance collapse issue of DARTS.
Moreover, it shows superior architecture search ability by finding the state-of-the-art ones on DARTS and NAS-Bench-201 search space.
The obtained architectures searched by our framework generalize well across CIFAR-10, CIFAR-100, and ImageNet, outperforming all previous DARTS variants.

\newpage
{\small
\bibliographystyle{ieee_fullname}
\bibliography{egbib}
}

\appendix
\section{Architecture Visualization}
\label{sec:arch_visual}
\begin{figure}[h]
  \centering
  \begin{subfigure}{0.4\linewidth}
    \centering
    \includegraphics[width=\linewidth]{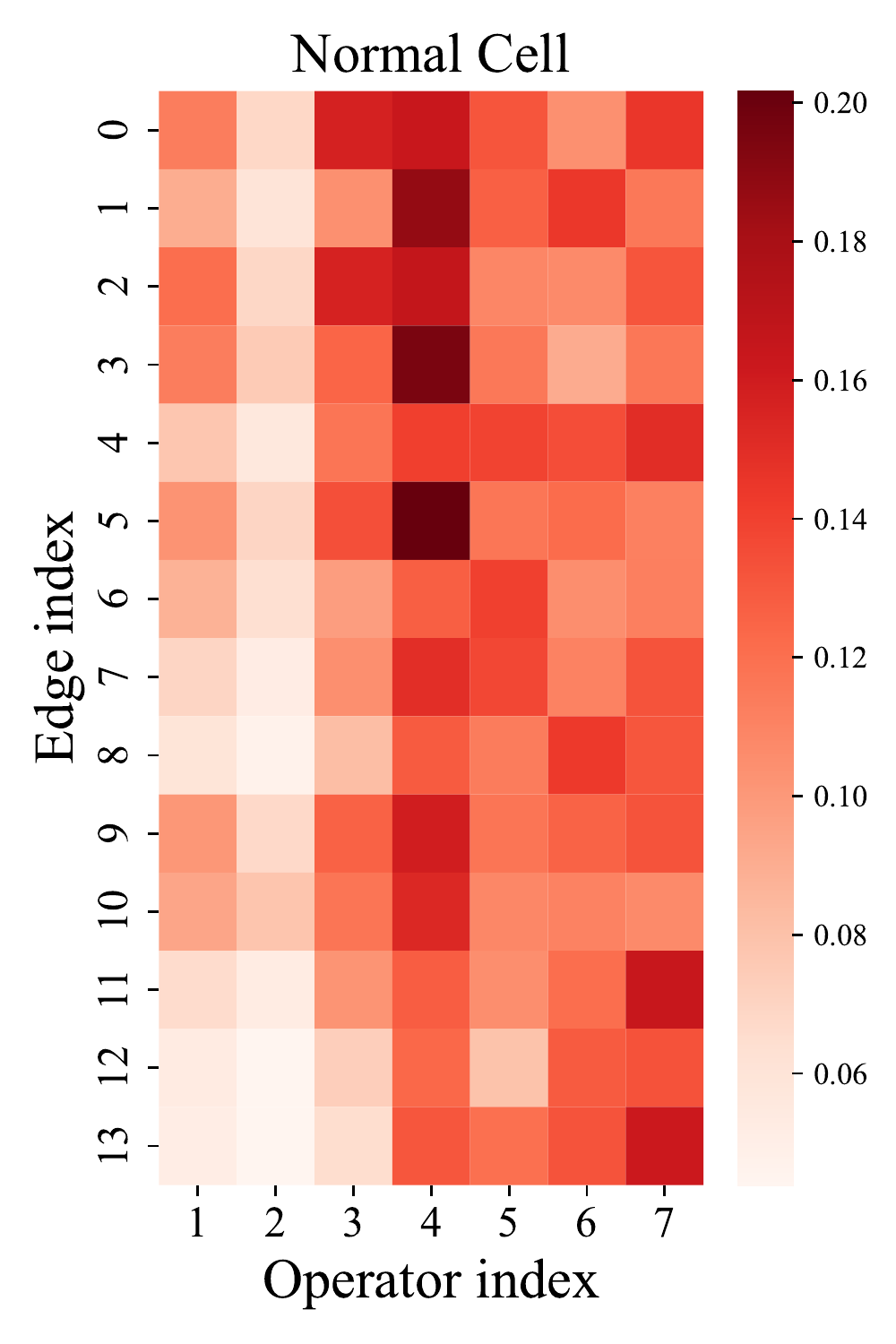}
  \caption{ours, total std=0.70}
  \end{subfigure}
  \begin{subfigure}{0.4\linewidth}
    \centering
    \includegraphics[width=\linewidth]{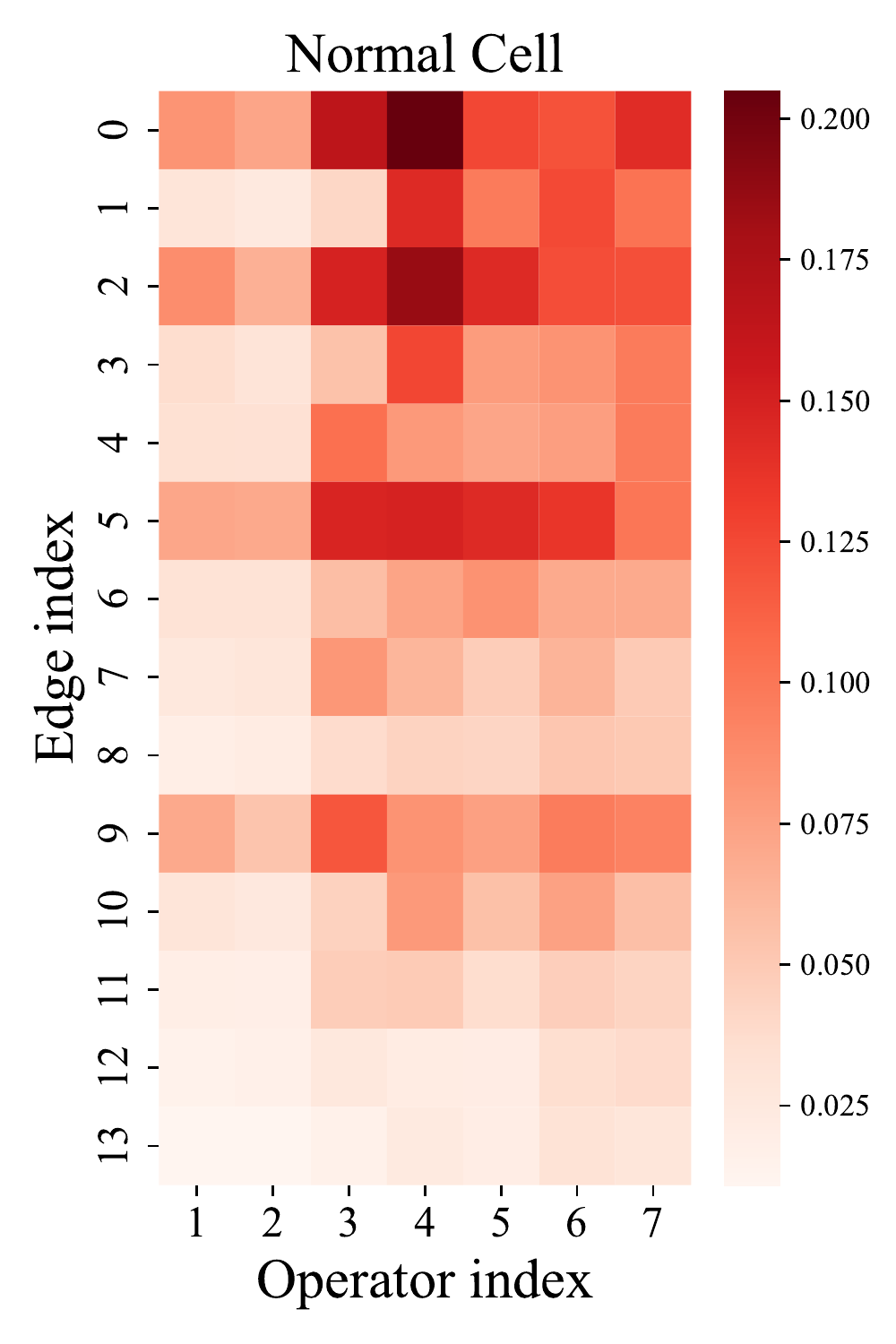}
  \caption{DARTS, total std=2.19}
  \end{subfigure}
  \caption{\textbf{The $\alpha$ distribution of normal cell learned by ours and DARTS on CIFAR-10.} 
    The operator indexes 1,2,3,4,5,6 denote 3$\times$3 max pooling, 3$\times$3 average pooling, skip connect, 3$\times$3 and 5$\times$5 separable convolution, 3$\times$3 and 5$\times$5 dilated convolution. 
    The total standard deviation is calculated by the sum of the standard deviation of all edges.
    The $\alpha$ values for each edge are normalized by softmax.
  }
  \vspace{-8pt}
  \label{fig:beta_distri}
\end{figure}

\begin{figure*}
  \centering
  
  \begin{subfigure}{0.33\linewidth}
    \begin{subfigure}{1\linewidth}
      \includegraphics[width=1\linewidth]{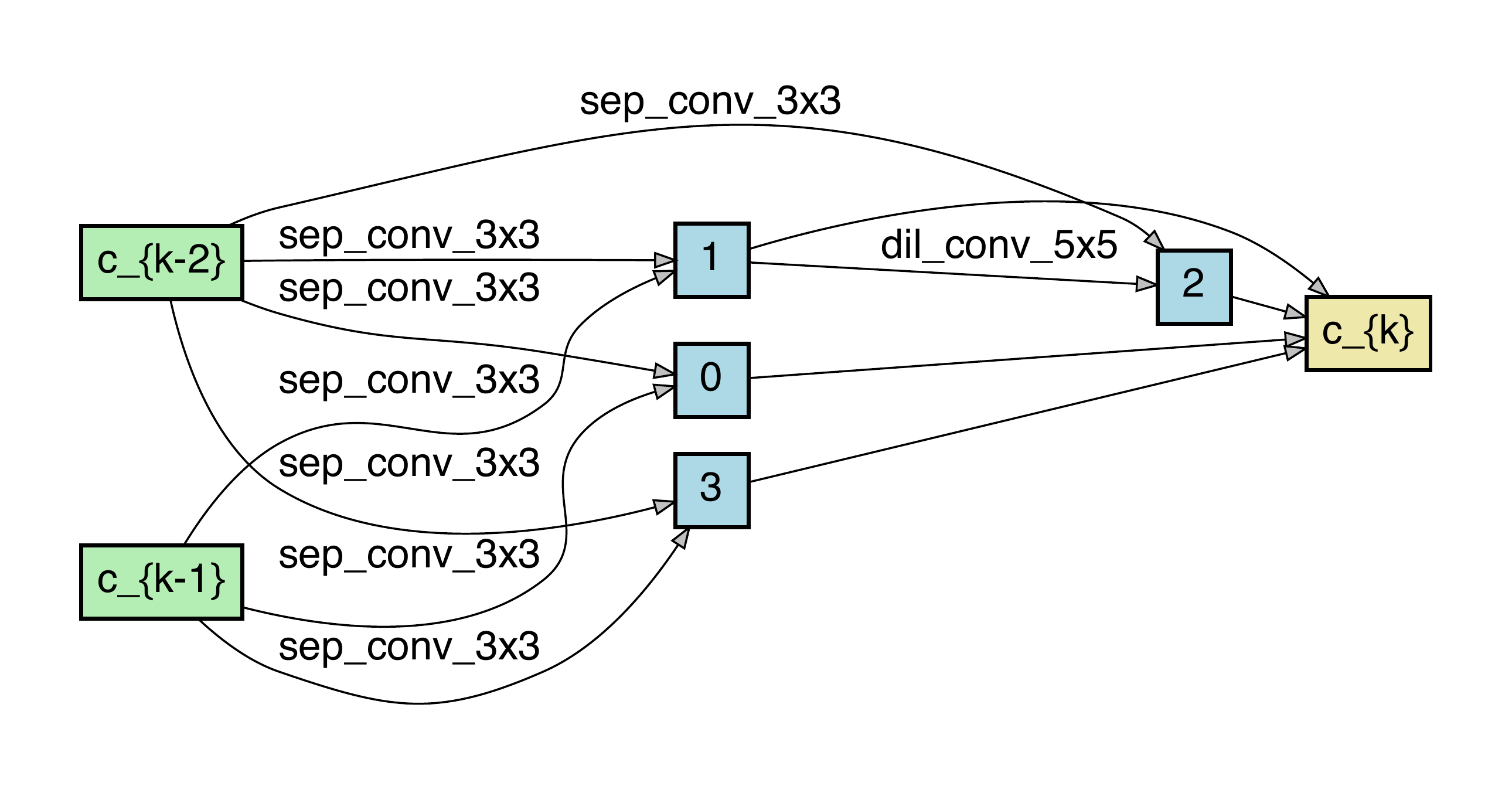}
    \end{subfigure}

    \begin{subfigure}{1\linewidth}
      \includegraphics[width=1\linewidth]{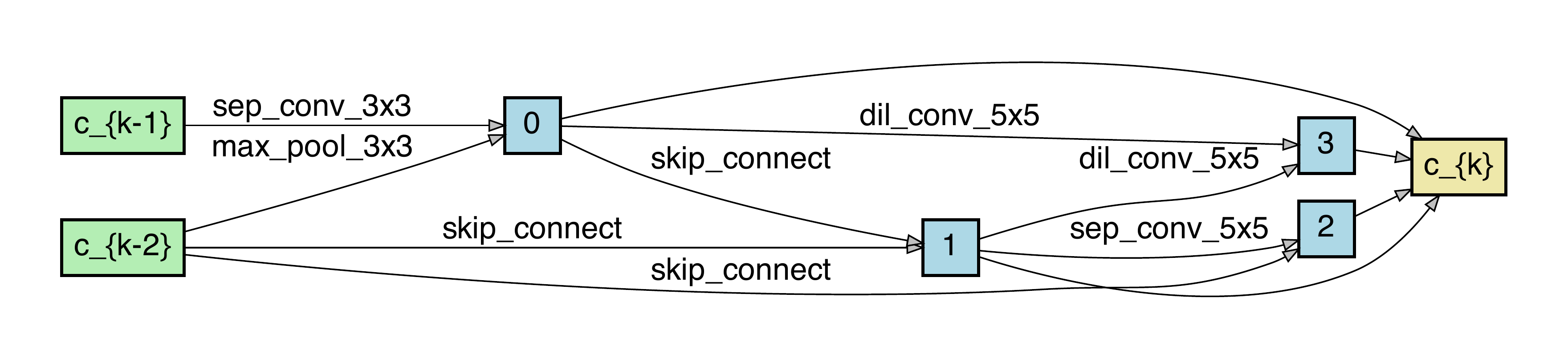}
    \end{subfigure}
  \caption{ImageNet Top-1 Accuracy 76.22\%}
  \end{subfigure}
  \begin{subfigure}{0.33\linewidth}
    \begin{subfigure}{1\linewidth}
      \includegraphics[width=1\linewidth]{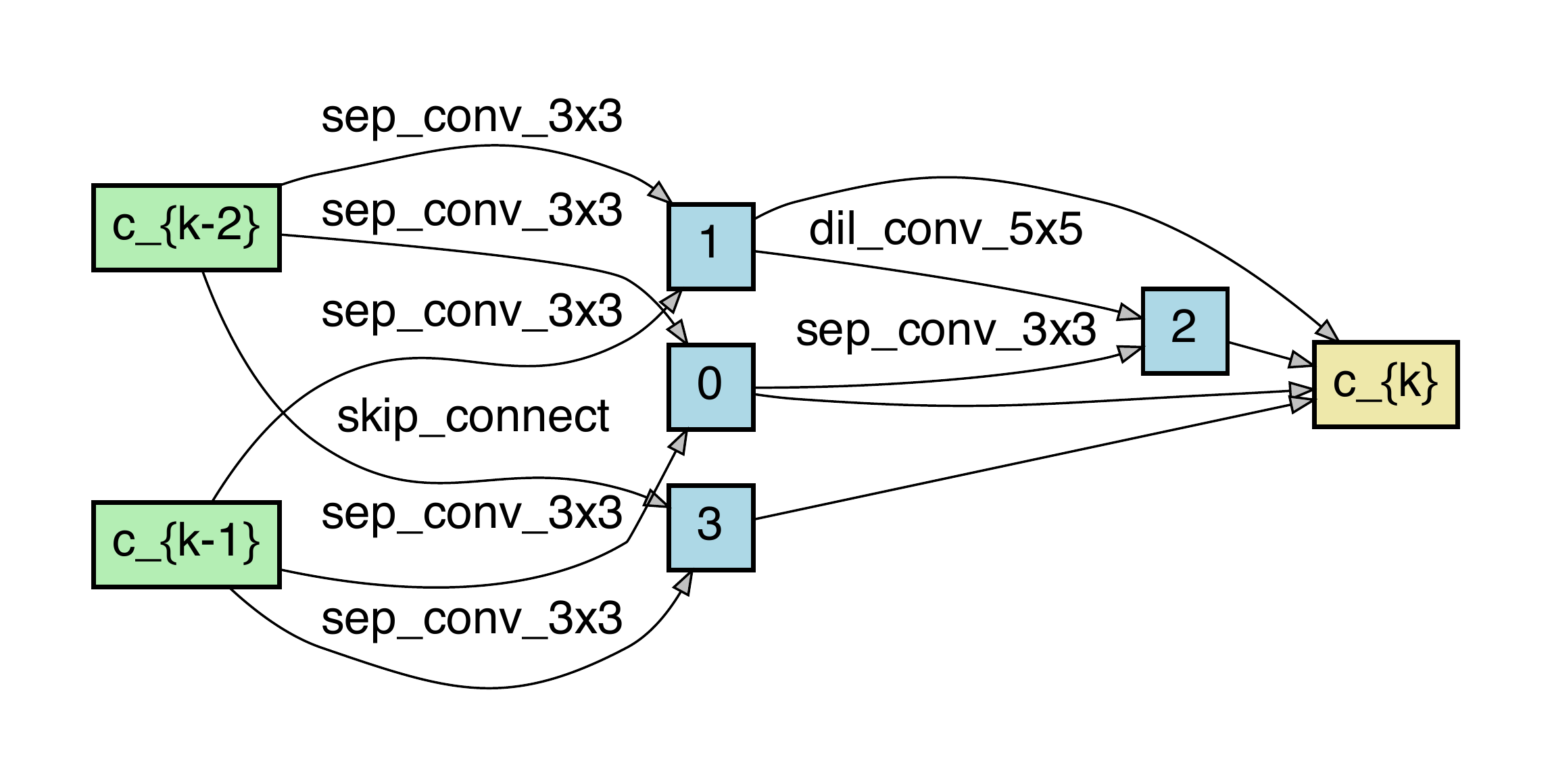}
    \end{subfigure}

    \begin{subfigure}{1\linewidth}
      \includegraphics[width=1\linewidth]{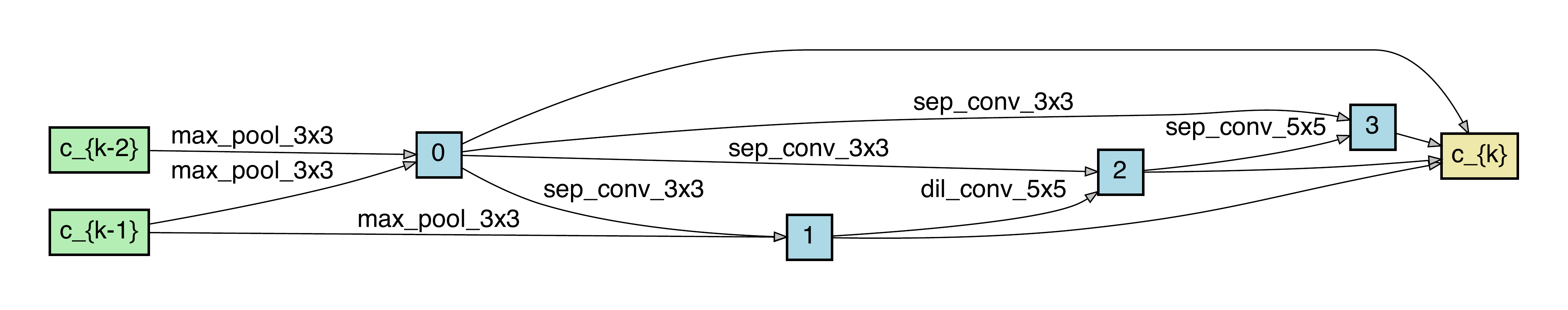}
    \end{subfigure}
  \caption{ImageNet Top-1 Accuracy 76.02\%}
  \end{subfigure}

  \begin{subfigure}{0.33\linewidth}
    \begin{subfigure}{1\linewidth}
      \includegraphics[width=1\linewidth]{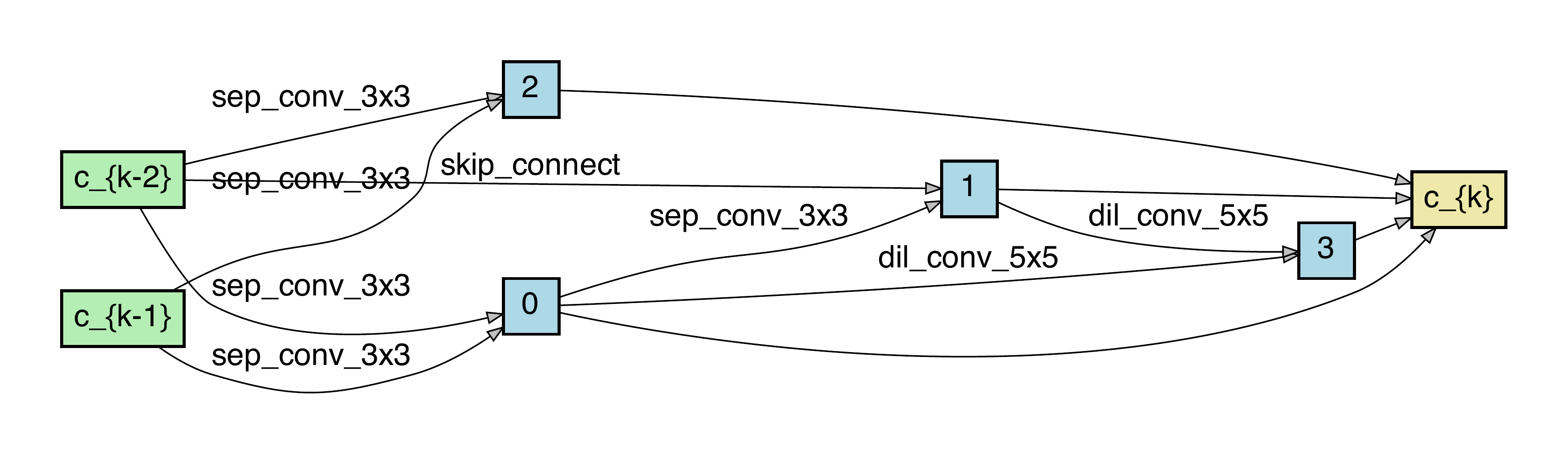}
    \end{subfigure}

    \begin{subfigure}{1\linewidth}
      \includegraphics[width=1\linewidth]{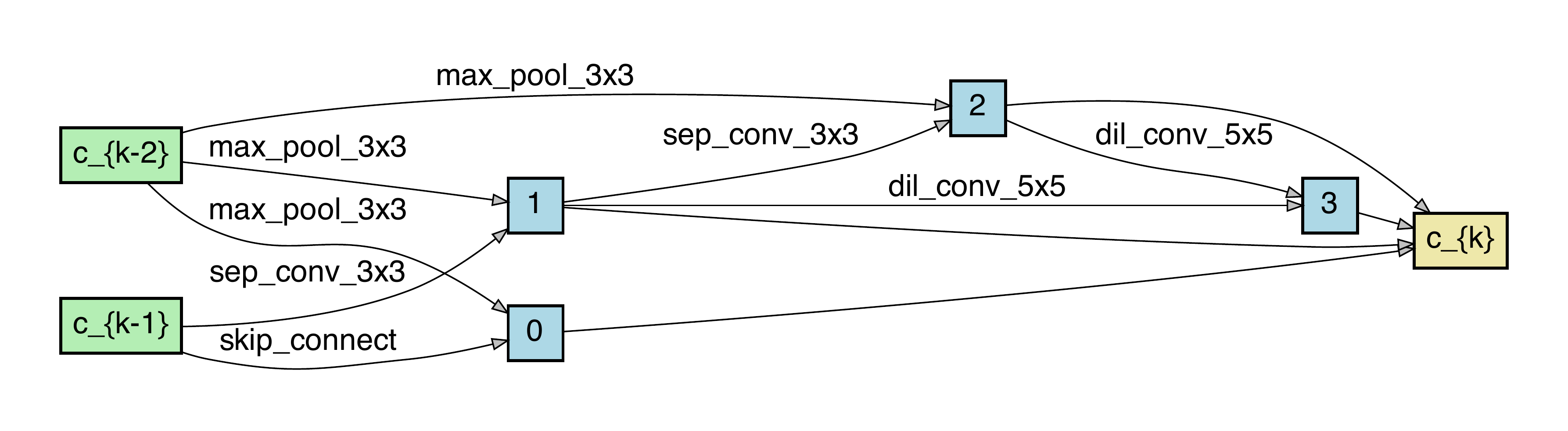}
    \end{subfigure}
  \caption{ImageNet Top-1 Accuracy 75.71\%}
  \end{subfigure}
  \begin{subfigure}{0.33\linewidth}
    \begin{subfigure}{1\linewidth}
      \includegraphics[width=1\linewidth]{figure/archs/normal_cls_maskv1c10seed9999r6p8.pdf}
    \end{subfigure}

    \begin{subfigure}{1\linewidth}
      \includegraphics[width=1\linewidth]{figure/archs/reduction_cls_maskv1c10seed9999r6p8.pdf}
    \end{subfigure}
  \caption{ImageNet Top-1 Accuracy 76.52\%}
  \end{subfigure}
  \caption{\textbf{Normal and Reduction cells discovered by MIM-DARTS on CIFAR-10.}}
  \label{fig:arch_vis_all_c10}
\end{figure*}

\begin{figure*}
  \centering
  
  \begin{subfigure}{0.33\linewidth}
    \begin{subfigure}{1\linewidth}
      \includegraphics[width=1\linewidth]{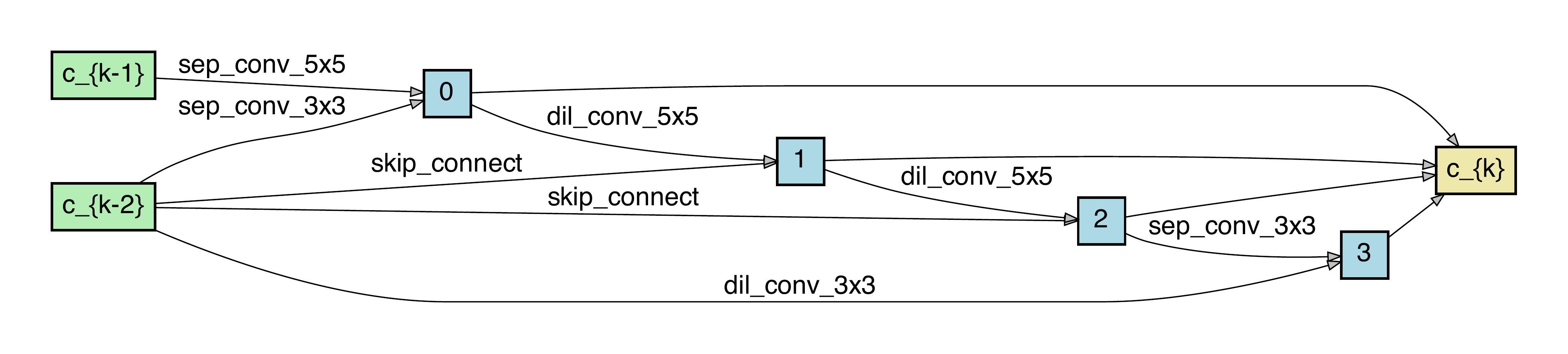}
    \end{subfigure}

    \begin{subfigure}{1\linewidth}
      \includegraphics[width=1\linewidth]{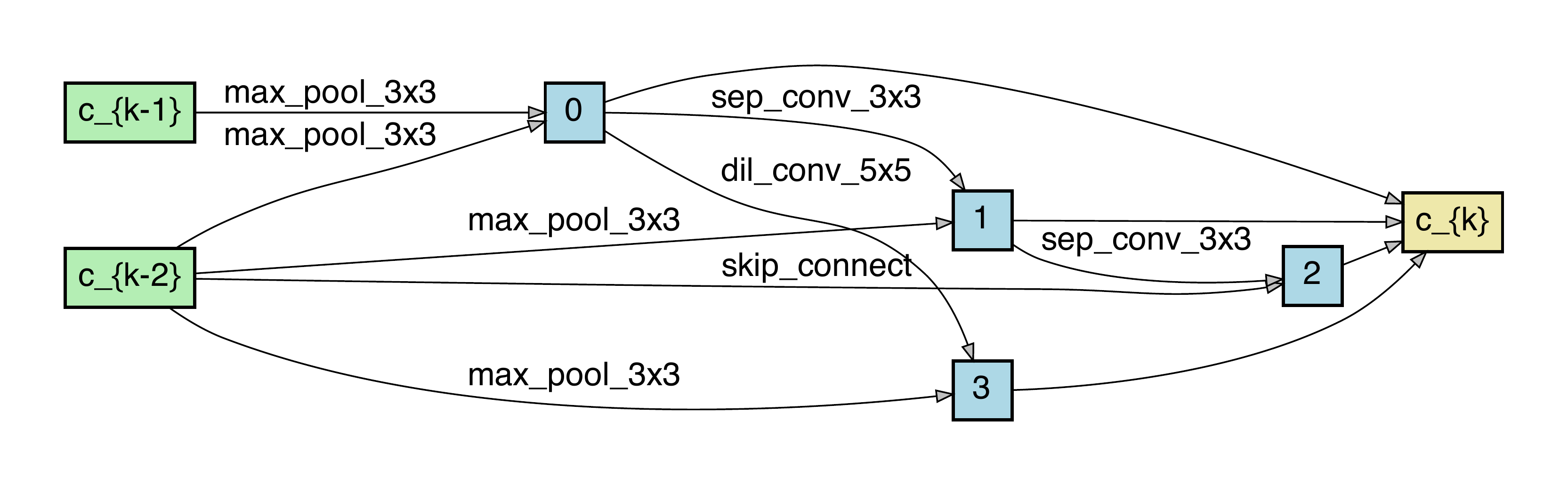}
    \end{subfigure}
  \caption{ImageNet Top-1 Accuracy 75.06\%}
  \end{subfigure}
  \begin{subfigure}{0.33\linewidth}
    \begin{subfigure}{1\linewidth}
      \includegraphics[width=1\linewidth]{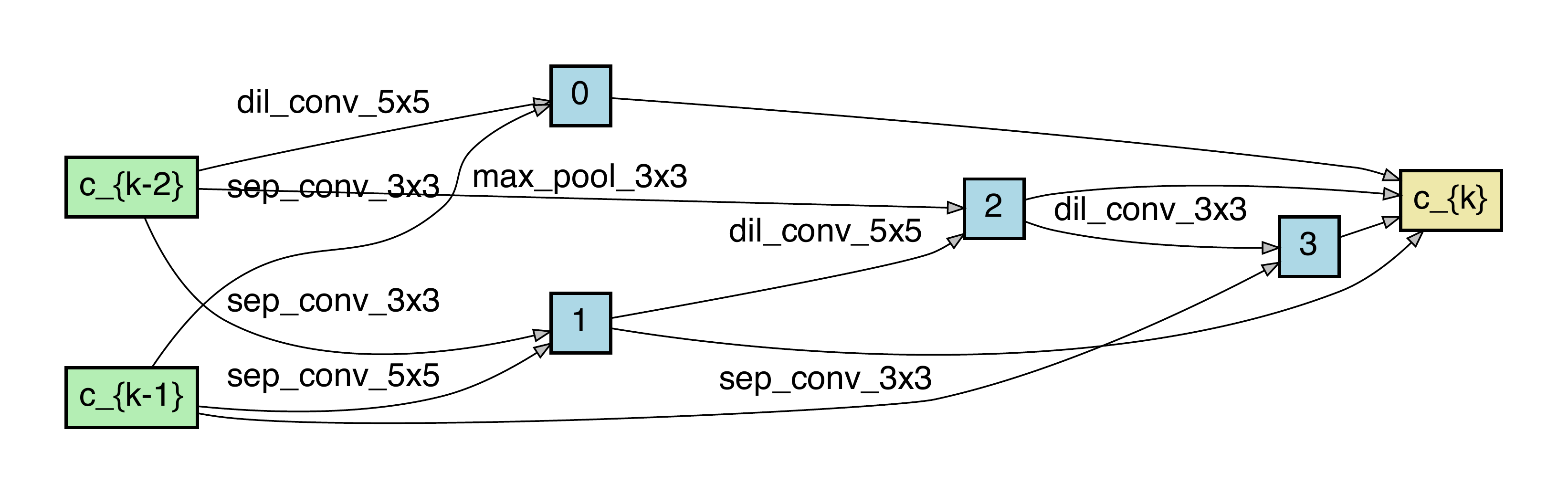}
    \end{subfigure}

    \begin{subfigure}{1\linewidth}
      \includegraphics[width=1\linewidth]{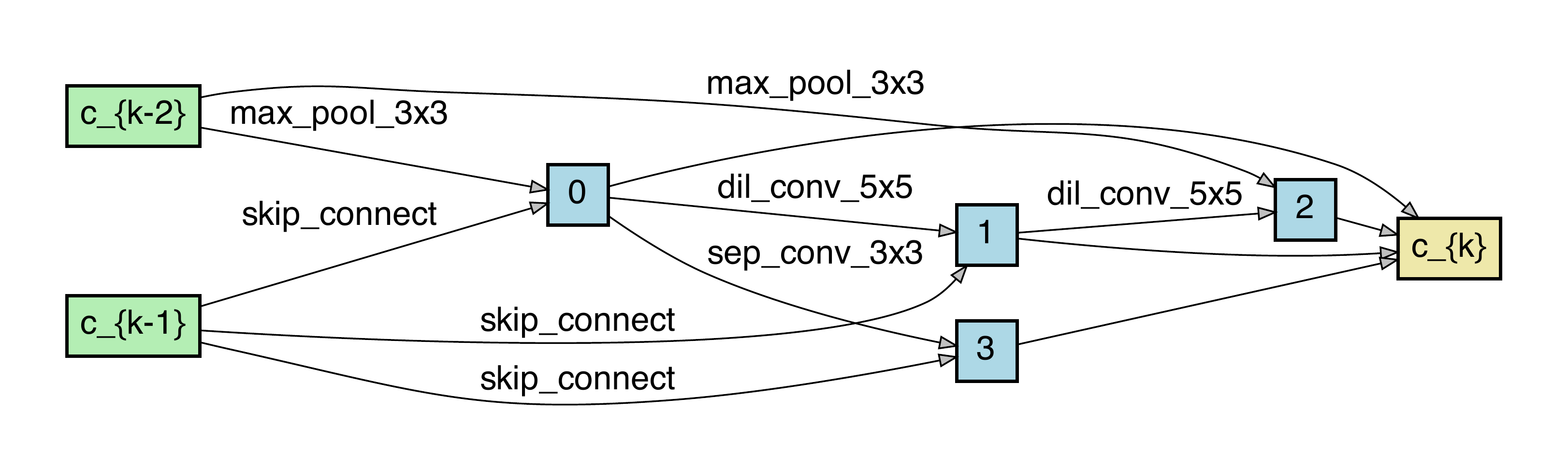}
    \end{subfigure}
  \caption{ImageNet Top-1 Accuracy 75.24\%}
  \end{subfigure}

  \begin{subfigure}{0.33\linewidth}
    \begin{subfigure}{1\linewidth}
      \includegraphics[width=1\linewidth]{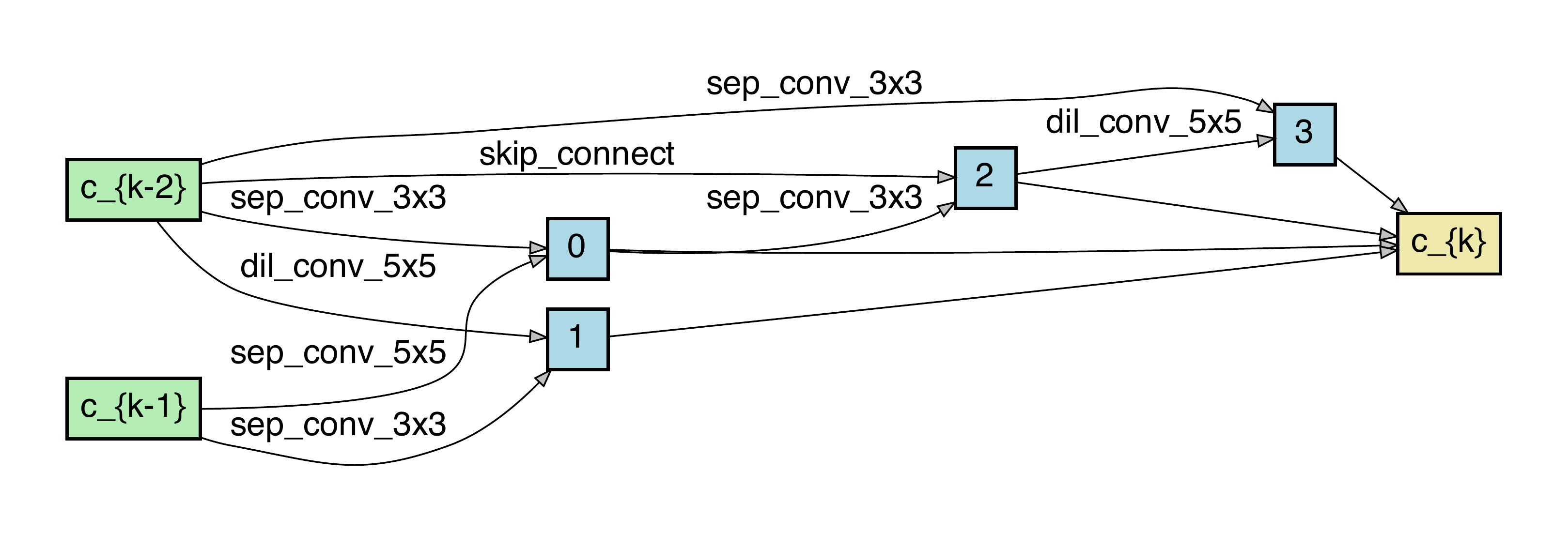}
    \end{subfigure}

    \begin{subfigure}{1\linewidth}
      \includegraphics[width=1\linewidth]{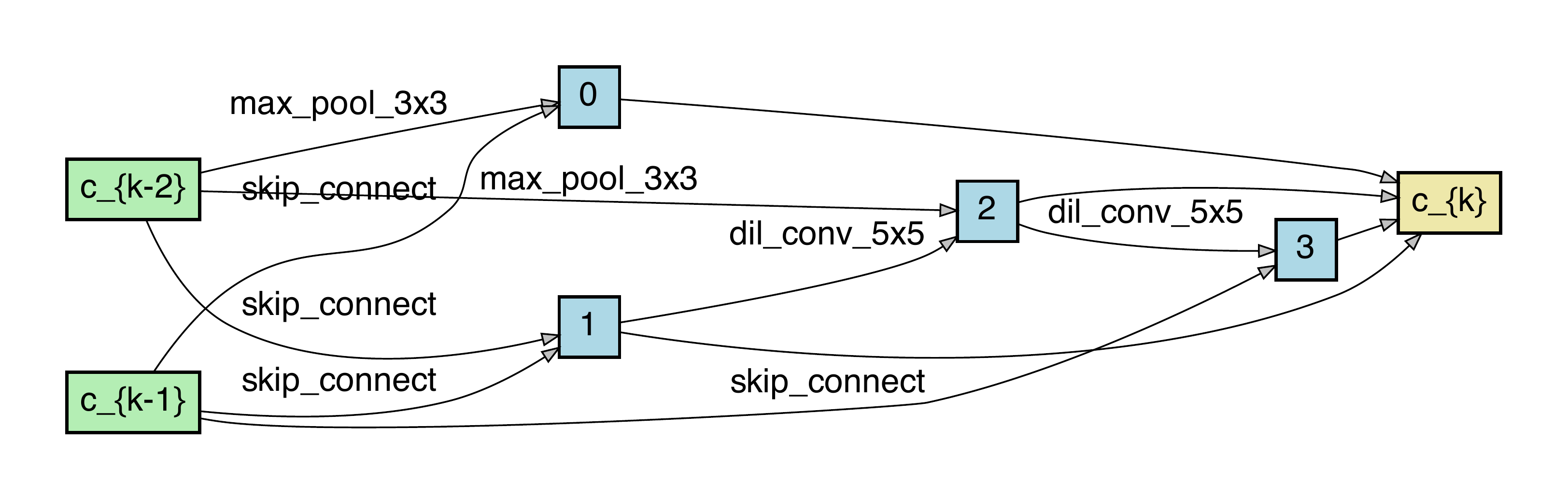}
    \end{subfigure}
  \caption{ImageNet Top-1 Accuracy 75.44\%}
  \end{subfigure}
  \begin{subfigure}{0.33\linewidth}
    \begin{subfigure}{1\linewidth}
      \includegraphics[width=1\linewidth]{figure/archs/normal_cls_maskv1c100seed9999r7p8.pdf}
    \end{subfigure}

    \begin{subfigure}{1\linewidth}
      \includegraphics[width=1\linewidth]{figure/archs/reduction_cls_maskv1c100seed9999r7p8.pdf}
    \end{subfigure}
  \caption{ImageNet Top-1 Accuracy 75.71\%}
  \end{subfigure}
  \caption{\textbf{Normal and Reduction cells discovered by MIM-DARTS on CIFAR-100.}}
  \label{fig:arch_vis_all_c100}
\end{figure*}

The visualization of the architectures searched from CIFAR-10 and CIFAR-100 are shown in \cref{fig:arch_vis_all_c10} and \cref{fig:arch_vis_all_c100}, respectively.
The ImageNet top-1 accuracy of each architecture is reported in the caption.
Further, we compare the $\alpha$ distributions of the obtained architectures from MIM-DARTS and the original DARTS~\cite{liu2018darts} as shown in \cref{fig:beta_distri}.
The accumulation of the standard deviation for all edges is reduced from 2.19 (the original DARTS) to 0.70 (MIM-DARTS), showing small variations of the $\alpha$ values.
This indicates that the learned architecture parameters finally converge to a flat area which is robust to the final discretization procedure for network operations selection as discussed in \cite{chen_2020_sdarts}.
Moreover, compared to the original DARTS, our method prefers parametric operators (operator index 4, 5, 6, 7) across all the edges, 
while the original DARTS shows this effect in only the first few edges.


\end{document}